\documentclass[twoside,11pt]{article}
\usepackage{jair, theapa, rawfonts}

\usepackage{amsmath,amsthm,amssymb}
\usepackage{mathtools}

\usepackage[table,xcdraw]{xcolor}
\usepackage{xspace}
\usepackage{graphicx}
\usepackage[figuresleft]{rotating}
\usepackage{lscape}
\usepackage{tabularx}
\usepackage{makecell}
\usepackage{paralist}

\usepackage[round]{natbib} 
\usepackage{todonotes}
\usepackage{bm}

\usepackage{comment}
\usepackage[breaklinks=true]{hyperref}
\usepackage{etoolbox}

\usepackage{algorithmic}
\usepackage{algorithm}

\hypersetup{colorlinks=true,linkcolor=blue,citecolor=blue,urlcolor=black}

\usepackage[english]{babel}
\addto\extrasenglish{

}

\renewcommand{\vec}[1]{\boldsymbol{#1}}
\newcommand*{\defeq}{\mathrel{\vcenter{\baselineskip0.5ex \lineskiplimit0pt
			\hbox{\footnotesize.}\hbox{\footnotesize.}}}%
	=}

\graphicspath{ {./images/} }

\newcommand*\rot{\rotatebox{90}}

\newcommand{\argmin}{\ensuremath{\mathrm{argmin}}}

\newcommand{\CluPaTra}{CluPaTra\xspace}

\newcommand{\CalibraLink}{\hyperref[Calibra]{Calibra}\xspace}
\newcommand{\ParamILSLink}{\hyperref[ParamILS]{ParamILS}\xspace}
\newcommand{\FRaceLink}{\hyperref[FRace]{F-Race}\xspace}
\newcommand{\SFRaceLink}{\hyperref[FRace]{Sampling F-Race}\xspace}
\newcommand{\IFRaceLink}{\hyperref[FRace]{Iterated F-Race}\xspace}
\newcommand{\IRaceLink}{\hyperref[FRace]{irace}\xspace}
\newcommand{\IRacecLink}{\hyperref[FRace]{irace(cap)}\xspace}
\newcommand{\GGALink}{\hyperref[par:GGA]{GGA}\xspace}
\newcommand{\PyDGGALink}{\hyperref[par:GGA]{PyDGGA}\xspace}
\newcommand{\HORALink}{\hyperref[HORA]{HORA}\xspace}
\newcommand{\ROARLink}{\hyperref[roar]{ROAR}\xspace}

\newcommand{\SMACLink}{\hyperref[SMAC]{SMAC}\xspace}
\newcommand{\WarmSMACLink}{\hyperref[SMAC]{WS SMAC}}
\newcommand{\SMACPSLink}{\hyperref[SMAC]{SMAC+PS}\xspace}
\newcommand{\DSMACLink}{\hyperref[SMAC]{D-SMAC}\xspace}
\newcommand{\MBGMLink}{\hyperref[GM]{MBGM}\xspace}
\newcommand{\BNTLink}{\hyperref[BNT]{BNT}\xspace}
\newcommand{\GGAPLink}{\hyperref[GGA++]{GGA++}\xspace}
\newcommand{\GPSLink}{\hyperref[GPS]{GPS}\xspace}
\newcommand{\REVACLink}{\hyperref[REVAC]{REVAC}\xspace}

\newcommand{\SPLink}{\hyperref[Structured]{SP}}
\newcommand{\LeapsAndBoundsLink}{\hyperref[LeapsAndBounds]{LeapsAndBounds}\xspace}
\newcommand{\SPwCLink}{\hyperref[SPwC]{SP$^*$ with Confidence}\xspace}
\newcommand{\CapsAndRunsLink}{\hyperref[CapsAndRuns]{CapsAndRuns}\xspace}
\newcommand{\ImpatientCapsAndRunsLink}{\hyperref[ImpatientCapsAndRuns]{ImpatientCAR}}

\newcommand{\ReACTLink}{\hyperref[ReACT]{ReACT}\xspace}
\newcommand{\ReACTRLink}{\hyperref[ReACT]{ReACTR}\xspace}
\newcommand{\CPPLLink}{\hyperref[CPPL]{CPPL}\xspace}

\newcommand{\ISACLink}{\hyperref[ISAC]{ISAC}\xspace}
\newcommand{\EISACLink}{\hyperref[ISAC]{EISAC}\xspace}
\newcommand{\ISACPLink}{\hyperref[ISAC]{ISAC++}\xspace}
\newcommand{\PCITLink}{\hyperref[PCIT]{PCIT}\xspace}
\newcommand{\CluPaTraLink}{\hyperref[CluPaTra]{CluPaTra}\xspace}
\newcommand{\SufTraLink}{\hyperref[CluPaTra]{SufTra}\xspace}
\newcommand{\FloTraLink}{\hyperref[CluPaTra]{FloTra}\xspace}
\newcommand{\HydraLink}{\hyperref[Hydra]{Hydra}\xspace}
\newcommand{\MATELink}{\hyperref[MATE]{MATE}\xspace}

\newcommand{\MOParamILSLink}{\hyperref[MO-ParamILS]{MO-ParamILS}\xspace}
\newcommand{\SraceLink}{\hyperref[Multi-ObjectiveRacing]{S-Race}\xspace}
\newcommand{\SPRINTRaceLink}{\hyperref[Multi-ObjectiveRacing]{SPRINTRace}\xspace}

\newcommand{\DACRLLink}{\hyperref[p:dac-rl]{DAC-RL}\xspace}
\newcommand{\HCRSLink}{\hyperref[p:HCRS]{HCRS}\xspace}

\newcommand{\A}{\mathcal{A}}
\newcommand{\I}{\mathcal{I}}
\newcommand{\Pd}{\mathcal{P}}
\newcommand{\E}{\mathbb{E}}

\newcommand{\M}{\mathcal{M}}

\makeatletter
\patchcmd{\NAT@citex}
  {\@citea\NAT@hyper@{%
     \NAT@nmfmt{\NAT@nm}%
     \hyper@natlinkbreak{\NAT@aysep\NAT@spacechar}{\@citeb\@extra@b@citeb}%
     \NAT@date}}
  {\@citea\NAT@nmfmt{\NAT@nm}%
   \NAT@aysep\NAT@spacechar\NAT@hyper@{\NAT@date}}{}{}

\patchcmd{\NAT@citex}
  {\@citea\NAT@hyper@{%
     \NAT@nmfmt{\NAT@nm}%
     \hyper@natlinkbreak{\NAT@spacechar\NAT@@open\if*#1*\else#1\NAT@spacechar\fi}%
       {\@citeb\@extra@b@citeb}%
     \NAT@date}}
  {\@citea\NAT@nmfmt{\NAT@nm}%
   \NAT@spacechar\NAT@@open\if*#1*\else#1\NAT@spacechar\fi\NAT@hyper@{\NAT@date}}
  {}{}
\makeatother

\jairheading{75}{2022}{425-487}{02/2022}{10/2022}
\ShortHeadings{A Survey of Methods for Automated Algorithm Configuration}
{Schede, Brandt, Tornede, Wever, Bengs, Hüllermeier \& Tierney}
\firstpageno{425}

\begin{document}
 
\title{A Survey of Methods for Automated \\ Algorithm Configuration}

\author{\name Elias Schede \email elias.schede@uni-bielefeld.de\\
       \addr Decision and Operation Technologies Group, \\ 
       Bielefeld University, Bielefeld, Germany
       \AND
       \name Jasmin Brandt \email jasmin.brandt@upb.de\\
       \name Alexander Tornede \email alexander.tornede@upb.de\\
       \addr Department of Computer Science, \\
       Paderborn University, Paderborn, Germany
       \AND
       \name Marcel Wever \email marcel.wever@ifi.lmu.de \\
        \addr Institute of Informatics, LMU Munich \& \\
        Munich Center for Machine Learning, Munich, Germany
       \AND
       \name Viktor Bengs \email viktor.bengs@ifi.lmu.de \\
        \addr Institute of Informatics, \\ LMU Munich,
       Munich, Germany
       \AND
       \name Eyke Hüllermeier \email eyke@lmu.de \\
       \addr Institute of Informatics, LMU Munich \& \\ 
       Munich Center for Machine Learning, Munich, Germany
       \AND
       \name Kevin Tierney \email kevin.tierney@uni-bielefeld.de \\
       \addr Decision and Operation Technologies Group, \\
       Bielefeld University, Bielefeld, Germany}

\maketitle

\begin{abstract}
Algorithm configuration (AC) is concerned with the automated search of the most suitable parameter configuration of a parametrized algorithm. There is currently a wide variety of AC problem variants and methods proposed in the literature. Existing reviews do not take into account all derivatives of the AC problem, nor do they offer a complete classification scheme. To this end, we introduce taxonomies to describe the AC problem and features of configuration methods, respectively. We review existing AC literature within the lens of our taxonomies, outline relevant design choices of configuration approaches, contrast methods and problem variants against each other, and describe the state of AC in industry. Finally, our review provides researchers and practitioners with a look at future research directions in the field of AC.
\end{abstract}

\section{Introduction}
Difficult computational problems must be regularly solved in many areas of industry and academia, such as constraint satisfaction problems, Boolean satisfiability problems (SAT), vehicle routing problems, finding a proper machine learning model for a given dataset, or computing highly complex simulations.
Algorithms that were developed to solve such problems usually have parameters that strongly influence the behavior of the respective algorithm and also, for example, the runtime that is required to solve problem instances or the quality of returned solutions.
In particular, different parameter values, also referred to as configurations in the following, are required for different sets of problem instances to achieve optimal results with respect to the running time or solution quality.
It is, therefore, crucial to adapt the configuration of an algorithm to the given data or specifics of the set of problem instances in question.
However, the configuration of algorithms, i.e., determining suitable parameter values, is a nontrivial and complex undertaking, since the algorithm must be actually executed for different configurations to observe the target metrics (e.g., runtime or solution quality).

The research field of algorithm configuration\footnotemark{} (AC) has emerged in response to this problem.
Especially within the last two decades, many approaches and problem variants have been proposed in this field.
Generally speaking, the approaches try to find effective configurations of algorithms as efficiently as possible and recommend them for new, unseen problem instances.
To this end, the approaches are typically given a training set of problem instances in an offline phase, which can be used as an input to run the algorithms, observe their performance, and (hopefully) generalize these observations to make good recommendations in production settings.
\footnotetext{In some works, the terms \textit{parameter tuning} and \textit{algorithm configuration} are used interchangeably~\citep{hoos2011automated}, however, we prefer \textit{algorithm configuration}, following the argumentation of~\cite{hutter2009paramils} that this term implies a more general configuration setting.}

To illustrate the benefits of automatically configuring algorithm parameters, we provide the circuit satisfiability problem as an example. It refers to a classic SAT problem in which the task is to find an assignment of values such that the output of a Boolean circuit evaluates to true~\citep{marques2008practical}. In a business application, many such circuits must be evaluated to check their feasibility in limited time. This requires an efficient SAT solver such as Glucose~\citep{audemard2009predicting} to provide assignments in a timely manner. Glucose exposes several parameters that influence the search for assignments and, therefore, the time needed to evaluate a circuit. 

Indeed, configurations of SAT solvers found by ParamILS~\citep{hutter2007automatic}, one of the first procedures to search for high-quality parameters in a structured way, achieve considerable speedups compared to default configurations of solvers. In particular, the configurations found by ParamILS reduce the arithmetic mean runtime for software verification instances for the SAT solver SPEAR from 787.1s to 1.5 seconds in the best case~\citep{hutter2007boosting}. More recently, PyDGGA~\citep{AnsoteguiPST21} reduced the solving time of the SAT solver SparrowToRiss~\citep{balint2013sparrow} on instances from the N-Rooks~\citep{lindauer2018warmstarting} dataset from 116 to 6.3 seconds. This shows that a practitioner will possibly achieve significant performance gains solving circuit assignments in the long run by only investing a limited amount of time into configuring Glucose for the specific task upfront.
The recommendation of configurations for unseen problem instances can be seen as a common property of all algorithm configuration problem variants, which differ in terms of their concrete setting specifications.
For example, there might only be a single (finite) categorical parameter to be configured, also referred to as algorithm selection~\citep{rice76}, ranging from just a handful up to thousands of choices \citep{tornedeWH20xas}. In the other extreme, the space of algorithm configurations may take into account many parameters and could comprise infinitely many possible configurations. Beyond that, the problem variants differ in several other aspects, such as the objective function to be optimized (runtime or solution quality), whether multiple objectives are to be considered simultaneously, whether training is performed offline or on the fly in an online setting, etc.
Depending on the specific properties of an AC problem, algorithm configuration approaches, also referred to as algorithm configurators, can be more or less suited, or cannot be applied at all.

In this paper, we provide an overview of different variants of both AC problems and algorithm configurators.
To this end, we propose two classification schemes: one for AC problems, and one for algorithm configurators.
Based on this, we structure and summarize the available literature and classify existing problem variants as well as approaches to AC.

The remainder of the paper is structured as follows. First, in Section~\ref{sec:problem-formulation}, we give a formal introduction into the setting of algorithm configuration, specify the scope of this survey, and discuss the relation between AC, AS and HPO.
In Section~\ref{sec:classification}, we present the classification schemes for AC problems and approaches that are used, in turn, to describe and compare existing algorithm configurators.
In Sections~\ref{sec:ModelFree} and \ref{sec:ModelBased}, we survey algorithm configuration methods grouped by the property of whether these methods are model-free or leverage a model respectively. Section~\ref{sec:Guarantees} deals with theoretical guarantees that can be obtained.
Different problem variants, such as realtime AC, instance-specific vs. feature-based, multi-objective, and dynamic AC are discussed in Sections~\ref{sec:Realtime} to \ref{sec:Dynamic}. Eventually, with the help of our classification schemes, we elaborate on appealing research directions in Section~\ref{ResearchDirections} and conclude this survey in Section~\ref{sec:conclusion}. A list of abbreviations used in this work can be found in Table \ref{table:abrevations}. In addition, we provide a list of useful software in Table \ref{table:usefullsoftware}. We note, however, that this list is by no means exhaustive; it is meant to provide an idea about available software at the time of publication.

\section{Problem Formulation}\label{sec:problem-formulation}

\subsection{Algorithm Configuration}\label{Problemsetting}

To describe the AC problem more formally, we introduce the following notation that is similar to \citet{hutter2009paramils}. Let $\mathcal{I}$ be a space of \textit{problem instances} over which a \textit{probability distribution} $\Pd$ is defined. 
Optional \textit{feature vectors} $\vec{f}_i \in \mathbb{R}^d$ with features $f_{i,1},...,f_{i,d}$ can be computed for problem instances $i \in \I$ coming from this space.
Furthermore, let $\A$ denote a parametrized \textit{target algorithm}, with parameters $p_1,...,p_k$ which may be of categorical or numerical nature. The (finite or infinite) domain of each parameter $p_i$ is denoted by $\Theta_{i}$ such that $\Theta \subseteq \Theta_{1} \times ...\times \Theta_{k}$ is the space of all feasible parameter combinations, i.e., the so-called \textit{configuration} or \textit{search space}. 
A concrete instantiation of the target algorithm $\A$ with a given configuration $\vec{\theta} \in \Theta$ is denoted by $\A_{\vec{\theta}}$. 
Furthermore, let $c: \mathcal{I} \times \Theta \rightarrow \mathbb{R}$ be a cost function from the space of cost functions $\mathcal{C}$, which quantifies the cost of running a given problem instance with a given configuration.\footnote{$\mathcal{C}$ merely limits the possible cost functions, but it is needed for formalizing an aggregation function.} Depending on the target algorithm, $c$ may be stochastic and contain noise.
Then, ideally, we would like to find the optimal configuration $\vec{\theta}^* \in \Theta$ defined as 

\begin{equation} \label{eq:1}
    \vec{\theta}^* \in \arg\min_{\vec{\theta} \in \Theta} \int\limits_{\mathcal{I}} c(i,\vec{\theta})\,d\Pd(i)\,\,\, .
\end{equation}

However, in practice, the distribution $\Pd$ over $\mathcal{I}$ is unknown, and thus we must resort to solving a proxy problem.\footnote{Formally, $c$ is also required to be integrable}  To this end, we are provided both a set of training instances $\mathcal{I}_\mathit{train} \subseteq \mathcal{I}$ and an aggregation function $m: \mathcal{C} \times 2^\mathcal{I} \times \Theta \rightarrow \mathbb{R}$. The aggregation function is usually the arithmetic mean or a variation thereof that is computed over the given problem instances by applying the given configuration to each of them and computing their cost. Similar to empirical risk minimization in machine learning, we then seek to find the configuration minimizing the aggregated costs across the training instances, i.e.,

\begin{equation} \label{eq:2}
    \widehat{\vec{\theta}} \in \arg\min_{\vec{\theta} \in \Theta} m(c, \mathcal{I}_\mathit{train}, \vec{\theta}).
\end{equation}

Informally, the problem can be expressed as: given a target algorithm with a set of parameters and a set of problem instances, find a configuration that yields good performance with respect to the cost measure across the set of problem instances. We will refer to automated approaches capable of finding such configurations as \textit{(algorithm) configurators}.

\paragraph{Configuration example}
To make AC more accessible and to illustrate the associated challenges, we use the previously mentioned circuit assignment problem with Glucose as a SAT solver. Glucose in version 4 has 41 parameters (with 13 binary and 28 continuous parameter domains) that constitute the configuration space $\Theta$. Table \ref{tab:Glucose} shows a subset of the parameters with their values, bounds and the effect they have on the search. Note that we have simplified the parameter descriptions. In fact, the complexity of understanding what the parameters actually do further emphasizes the need for automated AC, as in many cases practitioners may not fully understand the function of the parameters.

\begin{table}[h]
\centering
\footnotesize
\begin{tabular}{l|l|l|l}
Parameter & Type & Domain & Effect\\ 
\hline
Random Frequency (RF) & Float & [0, 1] & \makecell[l]{Probability of assigning a variable a random value \\ instead of using a heuristic.} \\
Conflict Factor (CF) & Float & [0, 1] & \makecell[l]{ Factor for comparing the state of the current restart \\ to the state of the search as a whole.} \\ 
Restart Queue (RQ) & Int & $\{10, 11, \ldots\}$ & \makecell[l]{Moving average window size over clause conflicts \\ for computing a score of the current state.} \\ 
Preprocessing (PR) & Cat. & \{on, off\} & \makecell[l]{ Activate or deactivate preprocessing of problem \\ instances before the main search.} \\

\end{tabular}
\caption{\label{tab:Glucose} Subset of Glucose parameters.}
\end{table}

Suitable configurations are very problem dependent and need to be chosen carefully for the application at hand. For instance, the \emph{Random Frequency} parameter of Glucose influences the exploration/exploitation tradeoff. A setting of $p_{RF}=1$ will lead to a random search. On the contrary, $p_{RF}=0$  will let Glucose only assign values based on its heuristic, which could lead to a local optimum. Depending on the search landscape of the circuits, the appropriate value for the random frequency, and all others, may vary. In addition to setting single parameter values, parameter interactions play an important role in algorithm performance. Consider the parameters \emph{Conflict Factor} and \emph{Restart Queue}, which both influence the restart policy~\citep{huang2007effect}, in which the search is restarted from the top of the search tree. To force Glucose to perform restarts more often, both $p_{CF}$ and $p_{RQ}$ need to be set to reasonable values~\citep{audemard2012refining}. That is, if not set properly, they may contradict each other and result in unintended or ineffective restart behavior.

While a clever practitioner could use his or her domain knowledge to set these parameters manually, usually it is not entirely clear exactly what settings (or parameter interactions) will lead to good performance.
Thus, instead of relying on domain knowledge and configuring and testing parameter values manually, a configurator as ParamILS may be used. A practitioner collects problem instances and selects a cost and aggregation function to solve \ref{eq:2}. In the context of our example, a specific circuit relates to one problem instance $i$ for which features $f_{i,d}$ can be computed as proposed by Kroer and Malitsky~(\citeyear{kroer2011feature}). The circuit can be assumed to come from a distribution $\Pd$ that specifies all possible circuits for the application and how often specific circuits are expected to occur. The practitioner approximates this distribution by gathering a limited amount of representative circuits from this distribution over time in a training set $\mathcal{I}_\mathit{train}$. Since the application at hand requires many circuits to be evaluated in a short amount of time, the cost function $c$ specifies runtime minimization. As an aggregation function $m$, the arithmetic mean over the training instances can be used for a given configuration. Based on these inputs, the algorithm configurator ParamILS is then tasked with finding a suitable configuration $\widehat{\vec{\theta}}$ for Glucose that can quickly solve the example circuits, and should also reduce the runtime for unseen circuits encountered in the future.

\subsection{Review Scope}\label{subsec:AS_and_HPO}

We select and review stand-alone AC methods that are suitable to solve the problem described in Section~\ref{Problemsetting}.
To identify relevant contributions in the literature we use the search terms \{\textit{Algorithm Configuration, Parameter Control, Parameter Tuning}\} combined with one of \{\textit{Automated, Automatic, Offline, Online, Realtime, Dynamic, Instance Specific, Per-Instance, Multiobjective, Feature Based, Optimal}\} as a prefix. We use Google Scholar as our search engine.   
We manually filter the search results to only include methods designed to automatically configure solvers without user interaction, and are able to handle large search spaces. Moreover, we omit articles related to algorithm selection (AS) and hyperparameter optimization (HPO). We define these areas in more detail later.
We follow citations forwards and backwards from all articles that we accept into the review to find articles that may lack the keywords above. We place such articles into the list of articles and filter them as previously discussed.
 
 \begin{figure}[tb]
  \centering
  \includegraphics[width=0.8\linewidth]{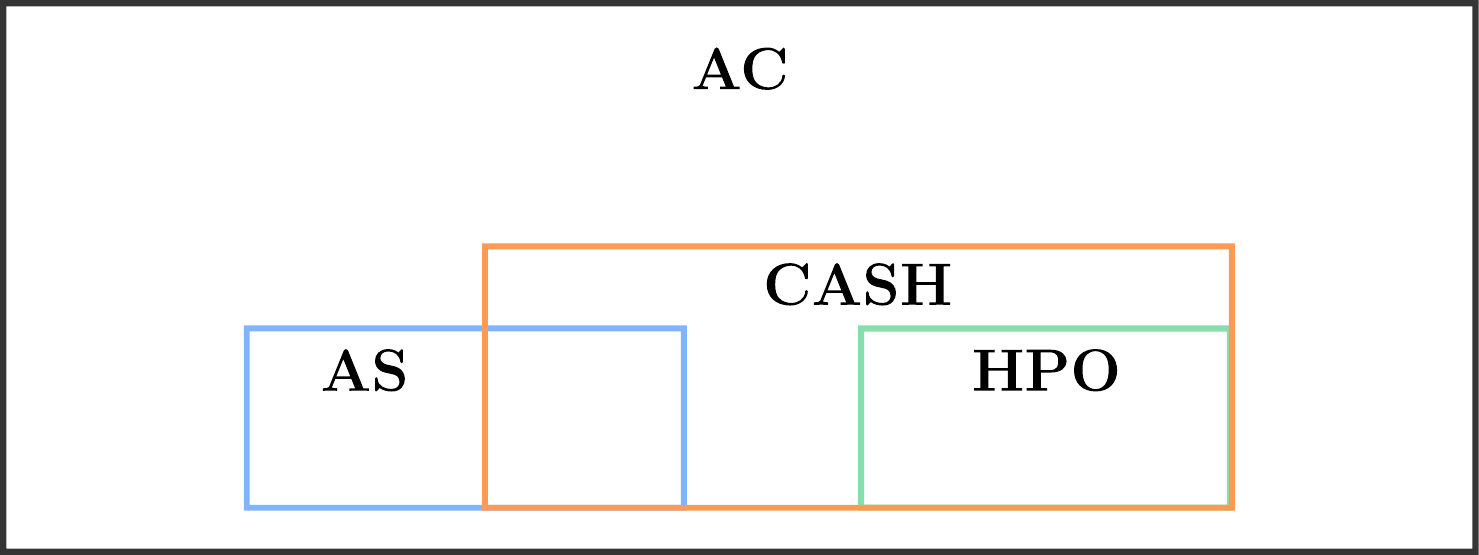}
  \caption{\label{fig:ac_AS_HPO}Illustration of the relationship between AC, AS, HPO and CASH.}
\end{figure}

\paragraph{Algorithm Selection (AS)}
AS is a sub problem of AC, however, we do not consider it in this work, as it has been considered in several reviews already~\citep{kerschke2019automated, kotthoff2016algorithm}. Thus, we now define what we mean by AS problems so that they can be filtered out of the review. AS can be seen as a special case of instance-specific AC (see Section \ref{sec:Instancespecific}), where the search space contains only one categorical parameter that models the target algorithm choice. In other words, the AS refers to learning how to configure that single categorical parameter, i.e. the algorithm choice, depending on the input instance. Thus it is severely restricted compared to AC. More specifically, the search space in AS is typically small, discrete and consists of a (static) set of algorithms (although new extensions exist that handle larger spaces~\citep{tornedeWH20xas}). The search space in AC, in general, is based on the parameters of one target algorithm, and thus, algorithm configurators need to be able to handle much larger, if not even infinite, search spaces~\citep{kerschke2019automated}. 

\paragraph{Hyperparameter Optimization (HPO)} Our review ignores HPO techniques in addition to AS, as these have also been considered in a number of reviews already~\citep{yu2020hyper,luo2016review,yang2020hyperparameter,hpoBischl}. Furthermore, before more clearly defining HPO, let us clarify the terminology around the words hyperparameter and parameter. In HPO, parameters that should be set by a user are referred to as hyperparameters, while in AC these are referred to as parameters. HPO refers to hyperparameters since machine learning models usually also contain parameters that are induced from data and are not considered by a configurator. In fact, it is this difference in terminology that leads us to one of the key differences between HPO and general AC, namely that AC methods focus on configuring target algorithms that solve instances of a dataset \emph{independently}, while HPO learns hyperparameters for target algorithms that train parameters on \emph{multiple} instances of a single dataset in tandem.

As HPO is a subset of the AC setting, HPO techniques can, in theory, be used to search for configurations in the general AC setting. In reality, this is seldom done because HPO methods ignore two key functionalities necessary for the general AC setting. First, HPO does not minimize algorithm runtime, which is often performed in general AC. HPO aims at optimizing a solution quality metric, such as predictive accuracy. Of course, AC settings exist where runtime is not a configuration objective, such as when configuring metaheuristics to find the best possible solution in a given time budget. Second, HPO techniques lack a problem instance selection mechanism. Specifically, one configuration in HPO is run on all the instances of one dataset and the result is observed by the configurator. Note that this set should be seen as a single AC problem instance, i.e., the term ``instance'' is used differently between the HPO and general AC communities. In AC, the configuration needs to be tested on a (sub)set of problem instances before the configurator can infer traits about its quality. Furthermore, HPO can be paired with algorithm selection, which is referred to as the combined algorithm selection and hyperparameter optimization problem (CASH)~\citep{DBLP:conf/kdd/ThorntonHHL13}. 

\paragraph{Automated Machine Learning (AutoML, CASH)} The HPO problem can be extended by an algorithm selection component. The task of selecting machine learning algorithms and simultaneously tuning their hyperparameters is formalized by \cite{DBLP:conf/kdd/ThorntonHHL13} as the combined algorithm selection and hyperparameter optimization (CASH) problem. Similar to HPO, the CASH problem can be classified as a sub-problem of algorithm configuration that is restricted to the domain of machine learning. Note that in the setting of AutoML, configurators typically face only a single AC problem instance in the form of a machine learning dataset. Due to this, we do not cover AutoML/CASH within this overview but instead refer the interested reader to comprehensive surveys \citep{elshawi2019automated,DBLP:journals/jair/ZollerH21,hutter2019automated}.

\section{Classification}\label{sec:classification}
We propose a classification scheme that separately covers 
\begin{inparaenum}[(1)]
\item the algorithm configuration setting and 
\item the configurator itself.
\end{inparaenum} More precisely, the \emph{problem view} describes the configuration setting a method tries to address. The \emph{problem view} consists of eight subcategories with an emphasis on the properties of the problem and the interaction between the configurator and target algorithm. The \emph{configurator view} consists of seven components that portray important aspects of a configurator. Both of these views are interconnected and complementary. Moreover, the configurator view can be interpreted as an answer to a problem setting, where specific features are added to the configurator as a response to the configuration setting. Existing classification schemes proposed in the literature until now~\citep{huang2019survey, eiben2011evolutionary, eiben2011parameter, stutzle2019automated, eryoldacsliterature} focus solely on the configurator and ignore the problem setting. The proposed taxonomy allows for a description and characterization of methods by aggregating information in tuples. The scheme (especially the \emph{problem view}) can also be used to derive new problem scenarios that have not been addressed before by combining different aspects in previously unseen ways. 
\begin{figure}[ht]
  \centering
  \includegraphics[width=0.8\textwidth]{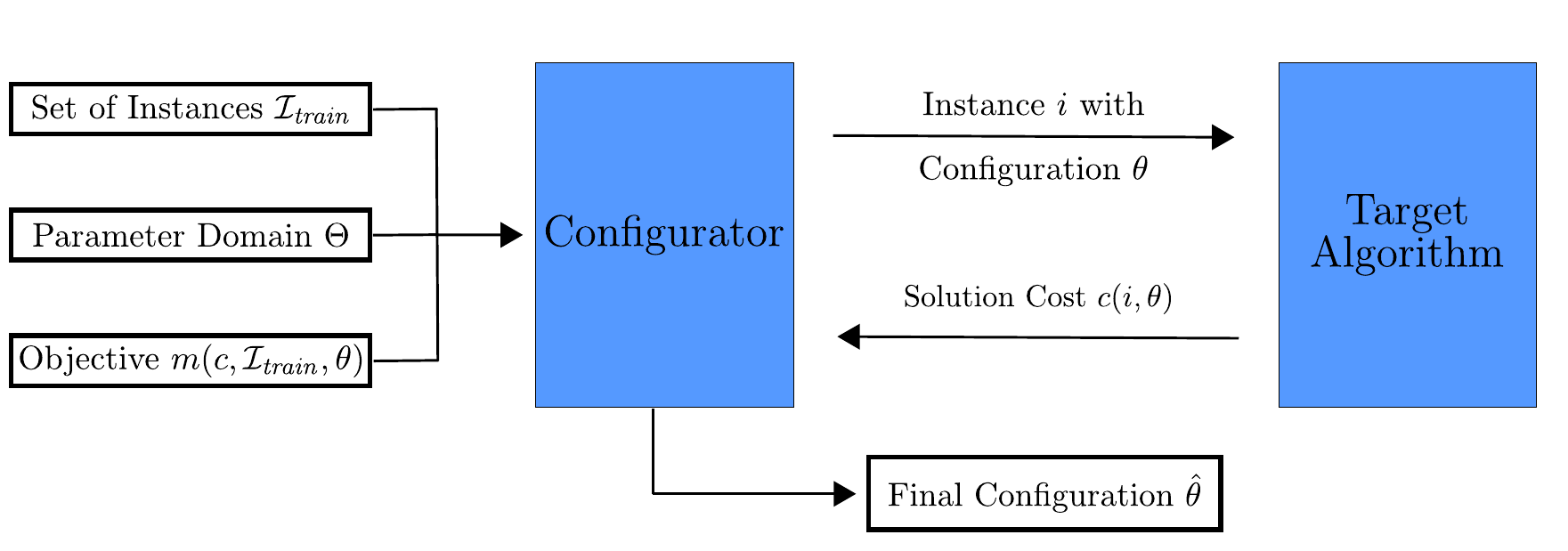}
  \caption{\label{fig:offline_configuration} Illustration of offline AC}
\end{figure}

\subsection{Problem View}
The components of the \emph{problem view} (Table~\ref{tab:Problem View_})  characterize a problem setting a configurator is meant for and therefore influence the configurator's design. Figure \ref{fig:offline_configuration} displays these interconnections and the communication between target algorithm and configurator, as well as the inputs a configurator receives. Note that, except for the \emph{objective function} and \emph{external runtime setting}, all other aspects are mutually exclusive, meaning that an unambiguous setting for a configurator exists. Furthermore, only the \emph{training setting} and \emph{configuration scope} are independent of the target algorithm. In the following, we elaborate on the dimensions and discuss their implications.

\begin{table}[h]
\begin{tabular}{l| |l|l|l}
\centering
\footnotesize
\bf Problem aspects & \multicolumn{3}{l}{\bf Options}\\
\hline
Training setting & Offline & Realtime &\\
Configuration scope & Set & Instance &\\
Search space & Small discrete & Large discrete & Infinite \\
Target algorithm objective type & Single-objective & Multi-objective &\\
Objective function$^*$ & Solving time & Accuracy & Memory Usage \\
Target algorithm observation time & During run & Post termination & \\
Configuration adjustment & Static & Dynamic &\\
External runtime setting$^*$ & Limited & Infinite &\\
\hline
\end{tabular}
\parbox[t]{\textwidth}{\footnotesize $^*$ Options not mutually exclusive}
\caption{\label{tab:Problem View_} The problem view classification scheme.}
\end{table}

\begin{description}
    \item[Training setting] Different training modes for configurators are possible: offline and realtime training or a combination of the two. In the offline setting, the configurator receives the set of training instances $\mathcal{I}_\mathit{train}$ as tuning begins, with which it searches for a suitable configuration. The setting is similar to the classic machine learning training setting, where multiple passes over the training set can be performed by the configurator and sufficient (possibly unlimited) time is available. Model-free and model-based offline methods are outlined in Sections~\ref{sec:ModelFree} and~\ref{sec:ModelBased}. In the realtime setting, the configurator receives a stream of problem instances, and it should solve each problem instance for the first time with a suitable configuration. While doing so, it can learn from solving the current problem instance to improve the solution time or quality of future configurations. In particular, the configurator is not trained up front, but sequentially during operation, and only one pass over the arriving data is possible. This setting is similar to the online learning setting in machine learning. Realtime methods can be found in Section~\ref{sec:Realtime}.
    \item[Configuration scope] The configurator can either be required to find a configuration for all problem instances in a set $\mathcal{I}_\mathit{train}$ or for individual problem instances. The former is referred to as a one-fits-all approach, i.e., a single configuration is derived that works well on average over a set of problem instances, whereas the latter requires configurations that are specifically derived for a problem instance. Determining instance specific configurations generally requires the configurator to take into account instance \emph{features}. Instance-specific configurators can be found in Section~\ref{sec:Instancespecific}.
    \item[Search space] The problem complexity is highly influenced by the \emph{search space} $\Theta$ spanned by the target algorithm parameters $p_1,...,p_k$. In particular, target algorithms may include a varying number of parameter types, which result in different search space sizes in which parameter transformations (e.g., logarithmic transformation) are possible~\citep{franzin2018effect}. 
    \item[Target algorithm objective type] The configurator may target a single objective or multiple objectives simultaneously.
    \item[Objective function] Different optimization goals exist, so designers must choose an \emph{objective function} $m(c, \mathcal{I}_\mathit{train}, \vec{\theta})$ for which a configurator optimizes. Usually, the configurator maximizes a solution quality metric, e.g., predictive accuracy, or minimizes runtime, but other metrics, such as minimizing memory usage, are also conceivable. Note that to handle target algorithm runs where no solution is found, penalty terms~\citep{eggensperger2019pitfalls} can be used. 
    \item[Target algorithm observation time] The time a configurator observes the cost $c(i,\vec{\theta})$ returned by the target algorithm can be either at termination of the target algorithm or during its execution.
    \item[Configuration adjustment] Target algorithms provide different opportunities for the configurator to adjust configurations, and this is orthogonal to the \emph{target algorithm observation time}. More precisely, the configuration $\vec{\theta}$ can either be adjusted only at the start of a run, where it then remains fixed, or the target algorithm may allow for dynamic adjustments of $\vec{\theta}$ during runtime. Configurators that are able to adjust configurations dynamically can be found in Section~\ref{sec:Dynamic}.
    \item[External runtime setting] The target algorithm may be influenced by an externally induced runtime cut off $\kappa_{max}$. Many AC settings set $\kappa_{max}$ explicitly, such as allowing a simulation to only run for a specified amount of time. The configurator can also limit $\kappa_{max}$, for example adaptively to reduce runtime, and we refer to this as the \emph{internal runtime setting} as part of the configurator view.
\end{description}

\subsection{Configurator View}
The \emph{configurator view} (Table~\ref{tab:Configurator View_}) characterizes algorithm configurators. The scheme does not cover concrete functionalities utilized by configurators such as intensification criteria or creation, selection and elimination of configurations. These functionalities are very difficult to characterize and classify, since for a single mechanism many options with only subtle differences may exist. In the following, we elaborate on relevant directions of the configurator view.

\begin{table}[h]
\begin{tabular}{l| |l|l|l}
\centering
\footnotesize
\bf Configurator aspect & \multicolumn{3}{l}{\bf Setting}\\
\hline
Solution quality guarantee & Heuristic & Proven & \\
Surrogate models & Model-free & Model-based & \\
Problem instance features & Featureless & Feature-based  & \\
Target algorithm execution & Sequential & Parallel & \\
Candidate output & Single configuration & Set configuration & Policy \\
Configurator objective & Single-objective  & Multi-objective  & \\
Internal runtime setting & Limited & Infinite  & \\
\hline
\end{tabular}
\caption{\label{tab:Configurator View_} The configurator view classification scheme.}
\end{table}

\begin{description}
    \item[Solution quality guarantee] Some recent AC methods offer theoretical guarantees regarding the quality of the configuration they return. However, most AC methods are heuristics that offer no proof about configuration quality. Methods that provide guarantees on the solution quality are discussed in Section~\ref{sec:Guarantees}.
    \item[Surrogate models] So-called model-based approaches incorporate surrogate models~\citep{hutter2005parameter, hutter2006performance, hutter2014algorithm} that are able to predict the outcome (e.g., runtime) of a target algorithm run. Such surrogate models (including empirical hardness models) are learned during training and can be used by the configurator to create new configurations, i.e., instead of trying a new configuration directly by running the target algorithm, the surrogate model is used to give a first estimate of the configuration quality. Because the surrogate models use the configuration as input, the models may also be referred to as models over configurations. For the sake of  of consistency we nevertheless will use the term model-based. Model-free and model-based approaches can be found in Section~\ref{sec:ModelFree} and Section~\ref{sec:ModelBased} respectively.
    \item[Problem instance features] For different problem types, features of problem instances can be computed and potentially be used by the configurator. In particular, feature-based approaches use a feature vector $\vec{f}_i$ with problem instance features $f_{i,1},...,f_{i,j}$. In the case of the SAT problem, an example for such a feature could be the ratio of the number of variables to the number of clauses~\citep{hutter2014algorithm}. The features are problem family dependent, meaning, e.g., mixed-integer programming problems (MILP) need to be described by different features than SAT problems. Features are commonly used in model-based approaches, e.g., within the surrogate model.
    \item[Target algorithm execution] A configurator may also be characterized by the way it executes and stops configuration runs of the target algorithm. In particular, configurators can run configurations in parallel on multiple cores or sequentially. 
    \item[Candidate output] Configurators may return one configuration $\hat{\vec{\theta}}$ or a set of configurations $\hat{\Theta}$, which can then be used by a decision maker or downstream process. In addition, it is possible for the configurator to return a policy that maps instances to configurations.
    \item[Configurator objective] While most configurators aim at optimizing one single objective \textbf{$m(\I,c(i,\vec{\theta}))$}, some configurators can handle multiple objectives $M:=(m_1,...,m_n)$. Note that the objective here refers to a combination of objective functions related to the configuration itself (e.g., a combination of solution quality and runtime), and not an output vector returned by the target algorithm (e.g., a target algorithm that returns a solution vector consisting of cost and environmental impact). Configurators that are able to handle multiple objective functions are discussed in Section~\ref{sec:MultiObjective}.
    \item[Internal runtime setting]In addition to the \emph{external runtime setting}, the configurator may decide to cancel a run of a target algorithm on a given configuration before the externally set runtime $\kappa_{max}$ is reached. We refer to the decision made by the configurator to either limit a run by capping or running it until the target algorithm terminates as the \emph{internal runtime setting}. The configurator may cancel runs according to some fixed cut off value or adaptively in relation to the runtime of other configurations. Although AC traditionally focused only on runtime capinng, recent work by~\citet{de2021capping} introduces capping mechanisms that can be used with quality metrics. Terminating target algorithm runs before they finish results in censored information for the configurator where only a lower bound for the costs $c(i,\vec{\theta})$ is observed~\citep{hutter2013bayesian, eggensperger2018efficient, eggensperger2020neural}. Previous work has shown that such right-censored data also needs to be handled properly in similar settings~\citep{run2survive,tornede2021ml4oas,hanselle2021algorithm}.
\end{description}
A full characterization of methods included in this review can be found in Table~\ref{tab:Problem View} and~\ref{tab:Configurator View}. Based on the classification scheme, existing configurators and ideas are discussed in the following. Starting with the most basic model-free approaches, methods are grouped by their most important characteristic or the main novelty they introduce.

\begin{table}[p]
\centering
\scriptsize

\begin{tabular}{l @{\hspace{0.8\tabcolsep}} l @{\hspace{0.8\tabcolsep}} l @{\hspace{0.8\tabcolsep}} l @{\hspace{0.8\tabcolsep}} l @{\hspace{0.8\tabcolsep}} l @{\hspace{0.8\tabcolsep}} l @{\hspace{0.8\tabcolsep}} l @{\hspace{0.8\tabcolsep}} l @{\hspace{0.8\tabcolsep}} l @{\hspace{0.8\tabcolsep}} l @{\hspace{0.8\tabcolsep}} l}
  Configurator &
  Reference &
  \rot{Training setting} &
  \rot{Configuration scope} &
  \rot{Target algorithm observation time} &
  \rot{Configuration adjustment} &
  \rot{Target algorithm objective type} &
  \rot{Search space} \\
  
\FRaceLink                  & {\tiny \citet{birattari2002racing}}           & O   & Set      & Pt & Sta  & Si  & Sd \\
\CalibraLink                & {\tiny \citet{adenso2006fine}}                & O   & Set      & Pt & Sta  & Si  & Sd \\
\HORALink                   & {\tiny \citet{barbosa2017heuristic}}          & O   & Set      & Pt & Sta  & Si  & Sd \\
\ParamILSLink               & {\tiny \citet{hutter2007automatic}}           & O   & Set      & Pt & Sta  & Si  & Ld\\
\HydraLink                  & {\tiny \citet{xu2010hydra}}                   & O   & Set      & Pt & Sta  & Si  & Ld\\
\BNTLink                    & {\tiny \citet{do2020automatic}}               & O   & Set      & Pt & Sta  & Si  & Ld\\
\REVACLink                  & {\tiny \citet{nannen2006method}}              & O   & Set      & Pt & Sta  & Si  & I       \\
\SFRaceLink                 & {\tiny \citet{balaprakash2007improvement}}    & O   & Set      & Pt & Sta  & Si  & I       \\
\IFRaceLink                 & {\tiny \citet{balaprakash2007improvement}}    & O   & Set      & Pt & Sta  & Si  & I       \\
\GGALink                    & {\tiny \citet{ansotegui2009gender}}           & O   & Set      & Pt & Sta  & Si  & I       \\
\PyDGGALink                 & {\tiny \citet{AnsoteguiPST21}}                & O   & Set      & Pt & Sta  & Si  & I       \\
\ROARLink                   & {\tiny \citet{hutter2011sequential}}          & O   & Set      & Pt & Sta  & Si  & I       \\
\SMACLink                   & {\tiny \citet{hutter2011sequential}}          & O   & Set      & Pt & Sta  & Si  & I       \\
\MBGMLink                   & {\tiny \citet{birattari2011learning}}         & O   & Set      & Pt & Sta  & Si  & I       \\
\DSMACLink                  & {\tiny \citet{hutter2012parallel}}            & O   & Set      & Pt & Sta  & Si  & I       \\
\GGAPLink                   & {\tiny \citet{ansotegui15gga++}}              & O   & Set      & Pt & Sta  & Si  & I       \\
\IRaceLink                  & {\tiny \citet{lopez2016irace}}                & O   & Set      & Pt & Sta  & Si  & I       \\
\IRacecLink                 & {\tiny \citet{caceres2017evaluating}}         & O   & Set      & Pt & Sta  & Si  & I       \\
\SPLink$^1$                 & {\tiny \citet{kleinberg2017efficiency}}       & O   & Set      & Pt & Sta  & Si  & I       \\
\LeapsAndBoundsLink         & {\tiny \citet{weisz2018leapsandbounds}}       & O   & Set      & Pt & Sta  & Si  & I       \\
\WarmSMACLink$^2$           & {\tiny \citet{lindauerH18warmstarting}}       & O   & Set      & Pt & Sta  & Si  & I       \\
\CapsAndRunsLink            & {\tiny \citet{weisz2019capsandruns}}          & O   & Set      & Pt & Sta  & Si  & I       \\
\SPwCLink                   & {\tiny \citet{kleinberg2019procrastinating}}  & O   & Set      & Pt & Sta  & Si  & I       \\
\GPSLink                    & {\tiny \citet{pushak2020golden}}              & O   & Set      & Pt & Sta  & Si  & I       \\
\ImpatientCapsAndRunsLink$^3$  & {\tiny \citet{weisz2020impatientcapsandruns}} & O   & Set      & Pt & Sta  & Si  & I       \\
\SMACPSLink                 & {\tiny \citet{anastacio2020model}}            & O   & Set      & Pt & Sta  & Si  & I       \\
\SraceLink                  & {\tiny \citet{zhang2013s}}                    & O   & Set      & Pt & Sta  & Mu   & Sd \\
\SPRINTRaceLink             & {\tiny \citet{zhang2015sprint}}               & O   & Set      & Pt & Sta  & Mu   & Sd \\
\MOParamILSLink             & {\tiny \citet{blot2016mo}}                    & O   & Set      & Pt & Sta  & Mu   & Ld\\
\HCRSLink                   & {\tiny \citet{DBLP:conf/aaai/AnsoteguiPST17}} & O   & Set      & Dr      & Dyn & Si  & I       \\
\DACRLLink                  & {\tiny \citet{biedenkapp2020dynamic}}         & O   & Set      & Dr      & Dyn & Si  & I       \\
\MATELink                   & {\tiny \citet{yafrani2020mate}}               & O   & In & Pt & Sta  & Si  & Sd \\
\CluPaTraLink               & {\tiny \citet{lau2011instance}}               & O   & In & Pt & Sta  & Si  & Ld\\
\FloTraLink                 & {\tiny \citet{lindawati2013flotra}}           & O   & In & Pt & Sta  & Si  & Ld\\
\SufTraLink                 & {\tiny \citet{yuan2013automated}}             & O   & In & Pt & Sta  & Si  & Ld\\
\ISACLink                   & {\tiny \citet{kadioglu2010isac}}              & O   & In & Pt & Sta  & Si  & I       \\
\EISACLink                  & {\tiny \citet{malitsky2014evolving}}          & O   & In & Pt & Sta  & Si  & I       \\
\ISACPLink                   & {\tiny \citet{ansotegui2016maxsat}}          & O   & In & Pt & Sta  & Si  & I       \\
\PCITLink                   & {\tiny \citet{DBLP:conf/aaai/LiuT019}}         & O   & In & Pt & Sta  & Si  & I       \\
\ReACTLink                  & {\tiny \citet{fitzgerald2014react}}           & Rt & In & Pt & Sta  & Si  & I       \\
\ReACTRLink                 & {\tiny \citet{fitzgerald2015reactr}}          & Rt & In & Pt & Sta  & Si  & I       \\
\CPPLLink                   & {\tiny \citet{el2020pool}}                    & Rt & In & Pt & Sta  & Si  & I \\

 \end{tabular}
\caption{\label{tab:Problem View} Configurators: problem view. $^1$Structured Procrastination, $^2$Warmstarting SMAC, $^3$CapsAndRuns. O: Offline, Rt: Real-time, In: Instance, Pt: Post termination, Dr: During run, Sta: Static, Dyn: Dynamic, Si: Single, Mu: Multi, Sd: Small discrete, Ld: Large discrete, I: Infinite.}
\end{table}

\begin{table}[p]
\centering
\scriptsize

\begin{tabular}{l @{\hspace{0.8\tabcolsep}} l @{\hspace{0.8\tabcolsep}} l @{\hspace{0.8\tabcolsep}} l @{\hspace{0.8\tabcolsep}} l @{\hspace{0.8\tabcolsep}} l @{\hspace{0.8\tabcolsep}} l @{\hspace{0.8\tabcolsep}} l @{\hspace{0.8\tabcolsep}} l @{\hspace{0.8\tabcolsep}} l}
  Configurator &
  Reference &
  \rot{Solution guarantee} &
  \rot{Surrogate model} &
  \rot{Instance features} &
  \rot{Algorithm execution} &
  \rot{Candidate output} &
  \rot{Configurator objective} &
  \rot{Internal runtime} &
  \rot{Distinguishing feature} \\

\FRaceLink                  & {\tiny \citet{birattari2002racing}}           & H & Mf  & Fl   & S  & Si & Si  & I & Racing \& F-test \\
\CalibraLink                & {\tiny\citet{adenso2006fine}}                & H & Mf  & Fl   & S  & Si & Si  & I & Taguchi design \& local search \\
\SFRaceLink                 & {\tiny\citet{balaprakash2007improvement}}    & H & Mf  & Fl   & S  & Si & Si  & I & F-Race \& sampling \\
\HORALink                   & {\tiny\citet{barbosa2017heuristic}}          & H & Mf  & Fl   & S  & Si  & Si   & I & Racing, DOE \& local search.\\
\ParamILSLink               & {\tiny\citet{hutter2007automatic}}           & H & Mf  & Fl   & S  & Si  & Si   & L  & Iterative local search\\
\ROARLink                   & {\tiny\citet{hutter2011sequential}}          & H & Mf  & Fl   & S  & Si  & Si   & L  & Random sampling \& racing\\
\SraceLink                  & {\tiny\citet{zhang2013s}}                    & H & Mf  & Fl   & S  & Set    & Mu   & I & Sign test \& racing\\
\SPRINTRaceLink             & {\tiny\citet{zhang2015sprint}}               & H & Mf  & Fl   & S  & Set    & Mu   & I & S  probability test \& racing \\
\MOParamILSLink             & {\tiny\citet{blot2016mo}}                    & H & Mf  & Fl   & S  & Set    & Mu   & L  & Iterative local search\\
\GGALink                    & {\tiny\citet{ansotegui2009gender}}           & H & Mf  & Fl   & Par   & Si  & Si   & L  & Genetic algorithm (GA) \& racing\\
\PyDGGALink                 & {\tiny\citet{AnsoteguiPST21}}                & H & Mf  & Fl   & Par   & Si  & Si   & L  & Genetic Distributed GGA\\
\ReACTLink                  & {\tiny\citet{fitzgerald2014react}}           & H & Mf  & Fl   & Par   & Si  & Si   & L  & Pool based \& racing\\
\ReACTRLink                 & {\tiny\citet{fitzgerald2015reactr}}          & H & Mf  & Fl   & Par   & Si  & Si   & L  & ReACT \& Trueskill\\
\REVACLink                  & {\tiny\citet{nannen2006method}}              & H & Mb & Fl   & S  & Si  & Si   & I & Distribution estimation \& GA\\
\IFRaceLink                 & {\tiny\citet{balaprakash2007improvement}}    & H & Mb & Fl   & S  & Si  & Si   & I & F-Race \& resampling \\
\MBGMLink                   & {\tiny\citet{birattari2011learning}}         & H & Mb & Fl   & S  & Si  & Si   & L  & Bayesian Networks \\
\BNTLink                    & {\tiny\citet{do2020automatic}}               & H & Mb & Fl   & S  & Si  & Si   & L  & Bayesian Networks \\
\IRaceLink                  & {\tiny\citet{lopez2016irace}}                & H & Mb & Fl   & Par   & Si  & Si   & I & F-Race \& elitist racing\\
\IRacecLink                & {\tiny\citet{caceres2017evaluating}}         & H & Mb & Fl   & Par   & Si  & Si   & L  & Irace \& capping\\
\GPSLink                    & {\tiny\citet{pushak2020golden}}              & H & Mb & Fl   & Par   & Si  & Si   & L  & Golden section search\\
\CluPaTraLink               & {\tiny\citet{lau2011instance}}               & H & Mb & Fb & S  & Si  & Si   & I & Clustering trajectories \& tuning \\
\FloTraLink                 & {\tiny\citet{lindawati2013flotra}}           & H & Mb & Fb & S  & Si  & Si   & I & CluPaTra \& suffix tree encoding \\
\SufTraLink                 & {\tiny\citet{yuan2013automated}}             & H & Mb & Fb & S  & Si  & Si   & I & CluPaTra \& graph trajectories \\
\SMACLink                   & {\tiny\citet{hutter2011sequential}}          & H & Mb & Fb & S  & Si  & Si   & L  & SMBO \\
\HCRSLink                  & {\tiny\citet{DBLP:conf/aaai/AnsoteguiPST17}}  & H & Mb  & Fb & S  & Si  & Si   & L  & GGA \&  logistic regression\\
\WarmSMACLink$^1$          & {\tiny\citet{lindauerH18warmstarting}}        & H & Mb & Fb & S  & Si  & Si   & L  & SMBO \& warmstart \\
\DACRLLink                  & {\tiny\citet{biedenkapp2020dynamic}}         & H & Mb  & Fb & S  & Si  & Si   & L  & Reinforcement learning (RL)\\
\SMACPSLink                 & {\tiny\citet{anastacio2020model}}            & H & Mb & Fb & S  & Si  & Si   & L  & SMBO \& probabilistic sampling\\
\HydraLink                  & {\tiny\citet{xu2010hydra}}                   & H & Mb & Fb & S  & Set    & Si   & L  & Boosting \\
\DSMACLink                  & {\tiny\citet{hutter2012parallel}}            & H & Mb & Fb & Par   & Si  & Si   & L & SMBO \& configuration queue\\
\ISACLink                   & {\tiny\citet{kadioglu2010isac}}              & H & Mb & Fb & Par   & Si  & Si   & L  & Clustering features \& Tuning \\
\EISACLink                  & {\tiny\citet{malitsky2014evolving}}          & H & Mb & Fb & Par   & Si  & Si   & L  & ISAC \& retraining \\
\ISACPLink                  & {\tiny\citet{ansotegui2016maxsat}}            & H & Mb & Fb & Par   & Si  & Si   & L  & ISAC \& AS \\
\PCITLink                   & {\tiny\citet{DBLP:conf/aaai/LiuT019}}            & H & Mb & Fb & Par   & Si  & Si   & L  & Clustering features \& Tuning\\
\GGAPLink                   & {\tiny\citet{ansotegui15gga++}}              & H & Mb & Fb & Par   & Si  & Si   & L  & GGA \& RF\\
\CPPLLink                   & {\tiny\citet{el2020pool}}                    & H & Mb & Fb & Par   & Si  & Si   & L  & ReACTR \& bandits\\
\MATELink                   & {\tiny\citet{yafrani2020mate}}               & H & Mb & Fb & Par   & Si  & Si   & L  & Genetic programming \& regression\\
\SPLink$^2$                 & {\tiny\citet{kleinberg2017efficiency}}       & P    & Mf  & Fl   & S  & Si  & Si   & L  & Emp. mean runtime queuing \\
\LeapsAndBoundsLink         & {\tiny\citet{weisz2018leapsandbounds}}       & P    & Mf  & Fl   & S  & Si  & Si   & L  & Phase-based Bernstein racing (BR) \\
\CapsAndRunsLink            & {\tiny\citet{weisz2019capsandruns}}           & P    & Mf  & Fl   & S  & Si  & Si   & L  & Timeout estimating \& BR  \\
\SPwCLink                   & {\tiny\citet{kleinberg2019procrastinating}}  & P    & Mf  & Fl   & S  & Si  & Si    & L  & Lower confidence bound SP \\
\ImpatientCapsAndRunsLink$^3$   & {\tiny\citet{weisz2020impatientcapsandruns}} & P    & Mf  & Fl   & S  & Si  & Si   & L  & CapsAndRuns with preprocessing \\

 \end{tabular}
\caption{\label{tab:Configurator View} Configurators: configurator view. $^1$Warmstarting SMAC, $^2$Structured Procrastination, $^3$CapsAndRuns. H: Heuristic, P: Proven, Mf: Model-free, Mb: Model-based, Fl: Featureless, Fb: Feature-based, S: Sequential, Par: Parallel, Si: Single, Mu: Multi, I: Infinite, L: Limited.}
\end{table}

\section{Model-free Methods}\label{sec:ModelFree}
\begin{figure}[h]
  \centering
  \includegraphics[width=0.95\textwidth]{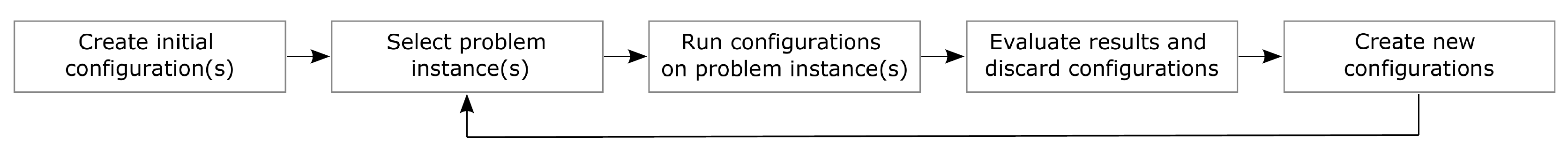}
  \caption{\label{fig:Search_process} Configuration search process of a configurator}
\end{figure}

We now describe model-free, offline AC approaches that solve the offline AC problem setting described in Section~\ref{Problemsetting}. Figure \ref{fig:Search_process} shows the general search process that is typically employed by configurators. In this context, there are three design choices that must be addressed by algorithm designers to solve the offline AC problem: 
\begin{inparaenum}[(1)]
\item the training instance sampling strategy,
\item the creation of configurations (initially and during search), and
\item the evaluation criteria for comparing configurations.
\end{inparaenum} 
Model-free configurators must run the target algorithm with a given configuration to observe its runtime or solution quality. Approaches that estimate the runtime or quality through models are described in Section~\ref{sec:ModelBased}. A variety of methods can be found in the literature that tackle each of the mentioned design choices differently, and we discuss them in the following.

\paragraph{Calibra}\label{Calibra}
One of the first approaches to systematically search for suitable parameter values is Calibra~\citep{adenso2006fine}. Calibra is a heuristic configurator that finds a configuration for a small discrete space of up to five parameters for a set of problem instances optimizing for solution quality. To do so, Calibra utilizes a combination of factorial experiments and local search. 

Factorial experiments are used to compare configurations, and a subsequent local search guides Calibra towards promising configurations. Initial configurations are created by a full factorial experimental design~\citep{fisher1937design} using the first and third quartile of the value ranges of the parameters as fixed starting points. This factorial design is the limiting component of Calibra, as factorial design does not scale. During the local search, Calibra creates new configurations by narrowing down value ranges from the previously tested configurations, successively focusing the search on promising regions of the parameter space. Taguchi $L_{9}(3^4)$ design~\citep{roy2010primer} is used to pairwise compare the resulting configurations. The value ranges of the parameters are narrowed in each iteration of a loop until an evaluation budget is exhausted. If a local optimum is found, the result is stored, and the local search is reinitialized.

\smallskip

We note that there are, in fact, some earlier AC approaches that use the design of experiments methodology (DOE). Some of the first approaches for AC based on DOE are given by~\citet{coy2001using,bartz2003experimental}; Ruiz and Maroto (\citeyear{ruiz2005comprehensive}) and Ridge and Kudenko (\citeyear{ridge2007tuning}), which, like Calibra, in practice do not scale to even moderately sized configuration spaces. However, DOE can be used to preprocess parameters to reduce the search space. Gunawan and Lau (\citeyear{gunawan2011fine}) propose using factorial experiments to rank parameters according to their importance and to set unimportant parameters to default values, while using a configurator to optimize the remaining, important ones. In later work,~\citet{gunawan2013real} propose to use fractional factorial experiments to decompose parameters into different categories before using the previous procedure to identify important parameters for each of the categories. Finally,~\citet{fallahi2014parameter} use DOE and clustering in an expert system that helps a user derive problem instance specific configurations.

\paragraph{ParamILS}\label{ParamILS}
ParamILS~\citep{hutter2007automatic} employs iterated local search (ILS)~\citep{lourencco2003iterated} to find high quality parameter configurations. It can handle discrete search spaces of large sizes with conditional parameters, and can optimize for either solution quality or runtime, making it one of the first general-purpose algorithm configurators proposed in the literature. 

ParamILS pairs a local search with two different types of configuration comparison procedures to determine which configuration to perturb next. More precisely, it uses a one-exchange neighborhood search in which one parameter is changed in each iteration. To avoid local optima, the search is restarted at random. The local search in ParamILS differs from Calibra, where the parameters are changed based on the bounds and mid-point of a narrowing value range. The resulting configurations are compared based on two procedures: BasicILS and FocusedILS. BasicILS compares two configurations on an equal number of problem instance runs, where the same problem instances (and random seeds) are used. A configuration is deemed superior if it provides a better objective value (e.g., mean runtime). In contrast, FocusedILS adjusts the number of evaluations for the configurations in question dynamically, utilizing an approach similar to racing. In particular, it uses a notion of dominance, in which a configuration is accepted over a competing configuration if it has been evaluated on more problem instances and has lower cost on those problem instances.

ParamILS suffers from a few shortcomings that are partly addressed in later work. The main drawback is the fact that it can only handle discrete parameter domains. In addition, it spends a significant amount of time running inferior configurations. To address this,~\citet{hutter2009paramils} introduce adaptive capping, which terminates configurations as soon as it becomes clear that they will not be better than the current incumbent. Cáceres and Stützle (\citeyear{caceres2017exploring}) adjust the one-exchange neighborhood principle of ParamILS to employ reduced variable neighborhood search (RVNS). This  allows for more than one parameter of a configuration to be changed in each local search iteration.

\paragraph{F-Race}\label{FRace}
Racing configurators run different parameter configurations against each other and discard inferior configurations based on non-parametric statistical tests. They are inspired by racing mechanisms from the machine learning literature~\citep{maron1993hoeffding, moore1994efficient}. \citet{birattari2002racing} and \citet{birattari2009tuning} introduce F-Race, the first racing procedure to address AC. F-Race starts by creating configurations to be raced through a full factorial design. Races are then performed sequentially, in which all (remaining) parameter configurations are evaluated on a problem instance. After all runs terminate, F-Race performs a Friedman test to determine whether the results obtained are significantly different from the results of the previous race. If this is the case, F-Race eliminates inferior configurations through pairwise tests. The remaining configurations are raced on the next problem instance until one remains, or a stopping criterion is met.

Extensions to F-Race, such as sampling F-Race, iterated F-Race (I/F-Race) and irace, improve upon the shortcomings of F-Race. Sampling F-Race~\citep{balaprakash2007improvement, birattari2010f} creates only a fraction of starting configurations by random sampling, limiting the exponential number of starting configurations. Iterated F-Race~\citep{balaprakash2007improvement, birattari2010f} not only shrinks a population of configurations sequentially, but is also able to replenish the set with new configurations between races. To create new configurations, a bivariate normal distribution over the parameter space is used, which in each iteration is parametrized by the previous race winner. irace~\citep{lopez2016irace} is based on iterated F-Race and adds soft-restarts and elitist racing. It replaces the one winner carried over between races by I/F-Race through an elitist set that is kept and updated over races. With the elitist set, configurations must prove viable over a sequence of problem instances, and not win a race just by chance. Soft restarts ensure that configurations in the set do not become too similar over multiple iterations. To this end, a distance-based diversity measure triggers a restart if the population becomes too similar. In this case, diversity is increased by sampling new configurations from a reset probability model. The introduction of adaptive capping~\citep{caceres2017experimental} prevents irace from spending time on evaluating unpromising configurations and makes it competitive for scenarios where costs are related to runtime. Similar to~\citet{hutter2011sequential} and~\citet{ansotegui15gga++}, \citet{caceres2017evaluating} add a random forest as surrogate to predict the potential costs of new configurations. However, this only leads to minor improvements. \citet{DBLP:conf/wsc/DijkMSG14} propose uRace for deterministic target algorithms.

The principles of racing can also be paired with other configuration approaches. \citet{yuan2013analysis} combine black box optimizers with a post-selection technique based on racing. Moreover, the optimizer is given a small computational budget for which it finds a set of promising configurations. The configurations are then raced to identify the best configuration among the identified set. Another approach that falls into this line of work proposes to pair evolutionary algorithms (EA) with racing~\citep{yuan2005hybrid, yuan2007combining}. The main shortcoming of using gradient-free optimization techniques for the AC problem is that they are limited to continuous parameters, and thus require discrete and categorical parameters to be rounded.

\paragraph{GGA}\label{par:GGA}
The gender-based genetic algorithm (GGA)~\citep{ansotegui2009gender} is a population-based approach in which configurations are encoded as individuals in one of two populations: competitive and non-competitive. It supports discrete and continuous parameters and can take parameter interactions supplied by the user into account. It is inherently parallel and optimizes for runtime or solution quality.

GGA combines a strong intensification procedure involving racing individuals from the competitive pool with diversification from the non-competitive population. In each generation of GGA, the competitive population is raced on a random subset of instances that increases linearly with each generation until either a target number of instances is reached or the entire training set is run in each generation. If the number of competitive members of the population is less than the number of CPU cores available on the machine, the race is split into multiple ``mini-tournaments'' with a number of individuals equal to the number of cores. When configuring for runtime, mini-tournaments are stopped as soon as a set number of individuals in the mini-tournament solve all the instances of the subset; usually one or two individuals. In this way, GGA saves significant runtime without needing to guess an instance timeout that could result in the instance not being solved at all. In the case of quality tuning, the mini-tournaments are run completely and the winners are selected based on the evaluation criteria set by the user.

Once the tournament phase is completed, GGA updates its population using custom recombination and mutation operators. In each generation, one third of the population is replaced and removed from the population. This third is replaced with new individuals that are created by combining a randomly selected winner from the competitive population with a randomly selected non-competitive individual. The recombination operator selects parameters from the competitive and non-competitive individual in a conservative fashion based on a dependency tree that represents how the parameters interact with each other. In subsequent work~\citep{ansotegui15gga++}, the recombination operator is extended to use a random forest surrogate. A mutation operator is applied to randomly change a small percentage of the parameters using Gaussian distributions for continuous and discrete parameters (with a mean equal to the current value of the parameter) and a uniform distribution for categorical parameters. The new individuals are inserted into the competitive or non-competitive population at random, and GGA moves to the next generation. GGA terminates if it runs out of time or a goal generation is reached and the average performance of the population stops improving. \citet{AnsoteguiPST21} present PyDGGA, an enhanced, distributed version of GGA, that is optimized for high-performance computing clusters. Further improvements as well as an approach to instance-specific configuration can be found  in~\citet{ansotegui2021boosting}.

\paragraph{HORA}\label{HORA}
The heuristic oriented racing algorithm (HORA)~\citep{barbosa2017heuristic,barbosa2017improving} combines elements from racing, DOE and neighborhood search. It can handle small search spaces and is able to optimize for either solution quality or runtime, however no capping is performed during races. HORA utilizes a response surface methodology to find an initial set of good configurations, which are used as seeds to iteratively create and race new configurations. More precisely, a few problem instances are selected at the start from the training set from which initial configurations are found through DOE. The resulting configurations are used as the basis for the subsequent search, in which they are altered using a search method that is similar to the one exchange neighborhood search of ParamILS. To compare the resulting configurations, races are conducted analogous to F-Race. The Friedman test is used to discard configurations. The cycle of generating new configurations and racing them continues until a stopping criterion is reached.

\paragraph{ROAR}\label{roar}
\citet{hutter2011sequential} introduce random online aggressive racing (ROAR), an offline configurator that samples new configurations uniformly at random and compares these to the current incumbent using an intensification mechanism that is similar to FocusedILS. In particular, ROAR requires as many target algorithm runs for a new configuration as for the current incumbent. The difference to FocusedILS is that for different, new configurations, the order of the problem instances they are tested on is not fixed. Moreover, problem instances and seeds to test new configurations are sampled at random from the set of evaluations the current incumbent was tested on. While being evaluated on these samples, new configurations are rejected as their cost is revealed to be clearly worse than the current incumbent. A configuration is accepted if its cost is better than the incumbent over all problem instances and seeds considered so far. This principle of aggressive racing is also later used by SMAC, with the addition of a predictive model.

\paragraph{GPS}\label{GPS}
Motivated by the belief that parameter landscapes may not be as complex as assumed (see also~\citep{harrison2019parameter}), Pushak and Hoos (\citeyear{pushak2020golden}) propose Golden Parameter Search (GPS), an offline procedure that exploits simple structures of parameter spaces by searching parameter configurations semi-independently in parallel. In particular, it works based on the assumptions that \begin{inparaenum}[(1)]
\item numeric target algorithm parameters are uni-modal and 
\item that there are no strong interactions between most of the parameters.
\end{inparaenum}

GPS combines elements of the golden section search algorithm~\citep{kiefer1953sequential} with common AC methods such as racing, capping and intensification. The search is based on parameter ranges (so-called brackets) that are believed to contain the best value for the parameter. The parameter values within the brackets are evaluated in parallel and independently of other brackets using a racing procedure with a permutation test to compare runs. After the target algorithm runs, brackets are expanded and shrunk using the golden section search algorithm, where ranges are updated by shifting values. In case of a bracket being updated, the other brackets are informed about the value adjustment of the respective parameter. In addition, GPS uses a bandit approach based on the relative importance of a parameter to prioritize target algorithm runs for more important parameters.

To speed up evaluation, GPS incorporates capping and intensification mechanisms while continuously populating a queue of configurations that should be run in the future. Initial evaluations are performed only on a small set of problem instances, and this set is gradually increased by sampling from the training set when parameter updates are performed. In addition, GPS modifies the bound multiplier of the capping mechanism to be adaptive itself by introducing a dependence on the size of the current problem instance set, leading to a less aggressive capping strategy compared to ParamILS or irace. Lastly, a queue is maintained that dynamically adjusts the amount of instance-configuration combinations that are run to ensure CPU resources are effectively utilized.

\section{Model-based Offline Methods}\label{sec:ModelBased}
In this section, we discuss model-based, offline AC approaches for solving the offline AC problem setting described in Section~\ref{Problemsetting}. Model-based methods leverage some form of learned model, such as a random forest, that captures knowledge about the performance of configurations as part of the optimization process. In the following, we provide a short background on sequential model-based optimization (SMBO) approaches, then describe the various model-based methods, starting with random forest methods and followed by Bayesian network based methods.

\subsection{Sequential Model-based Optimization}\label{subsec:model-based-ac:smbo}
SMBO approaches model the AC problem as a black box optimization problem. Formally, given a function $g: \Theta \rightarrow \mathbb{R}$ with input domain $\Theta \subset \mathbb{R}^d$, the goal is to find the point optimizing the function $g$:
\begin{equation}
    \vec{\theta}^* \defeq \arg\min_{\vec{\theta} \in \Theta} g(\vec{\theta}) \,\,\, .
\end{equation} 

The function $g$ is referred to as a black box (function) since very few assumptions are made about its structure except for the ability to perform evaluations given a point as input. When defining $\Theta$ as the configuration space and $g$ as the performance of the algorithm to configure, the mapping from AC to black box optimization appears straight forward, however a key challenge is deciding which instances from $\mathcal{I}$ to run with a given configuration and how to integrate the incomplete information about the cost into the model.

SMBO approaches iteratively refine a surrogate model $\widehat{g}: \Theta \rightarrow \mathbb{R}$ with the goal of approximating the original $g$ as best possible based on point-wise evaluations of $g$. To this end, they are powered by two main concepts, first and foremost the aforementioned surrogate model used to approximate $g$ and a so-called acquisition function $a: \Theta \times \widehat{\mathcal{G}} \rightarrow 2^\Theta$, where $\widehat{\mathcal{G}}$ is the space of all possible surrogate models $\widehat{g}$. In short, SMBO approaches iteratively leverage the acquisition function $a$ to generate a set of configurations to evaluate based on the current approximation of $g$, i.e., the surrogate model $\widehat{g}$. The evaluations of the configurations on $g$ are provided to the surrogate model to improve its approximation quality and, thus, obtain a better idea of the response surface of $g$. 

One of the assumptions to be imposed on $g$, which is particularly important for the problem of AC, is whether $g$ is a deterministic or stochastic function. Often, in AC we consider randomized algorithms, meaning that $g$ is stochastic, i.e., the evaluations of $g$ contain \textit{noise}. Thus, multiple evaluations at the same point can result in different outcomes. Furthermore, the previously mentioned issue of sampling $\mathcal{I}$ leads to a noisy estimation of each point, as evaluating all instances is usually too expensive until later stages of configuration.

We give a more detailed description of the SMBO framework for AC based on the pseudocode presented in Algorithm \ref{alg:smbo} in the following. At the start of the optimization, a so-called initial design strategy is used to generate a set of initial configurations to perform an initial training of the surrogate model $\widehat{g}$. For a further overview of these strategies, we refer to \citep{santner2003the}. Once the surrogate model is fit on these points, the main loop starts and the following steps are repeated until a stopping criterion is met. First, the acquisition function leverages the current surrogate model $\widehat{g}$ to generate a set of configurations that $g$ should be evaluated on. The set of configurations returned by the acquisition function is then passed on to the intensification procedure, which decides how many evaluations are performed per configuration provided. Lastly, the surrogate model is updated based on the configurations chosen by the acquisition function and the corresponding evaluation results provided by the intensification strategy.

\begin{algorithm}[ht]
    \caption{SMBO(configuration space $\Theta$, $\mathit{initial\_design}$, \textit{black box function} g,  \\ \textit{surrogate model} $\widehat{g}$, \textit{acquisition function} $a$)}
    \label{alg:smbo}
    \footnotesize
	\begin{algorithmic}[1]
	    \STATE $\widehat{\vec{\theta}}^* \gets \mathit{random}(\Theta)$
	    \STATE $X_\mathit{init} \gets \mathit{initial\_design}()$
	    \STATE $Y_\mathit{init} \gets g(X_\mathit{init})$
	    \STATE Fit $\widehat{g}$ to $(X_\mathit{init}, Y_\mathit{init})$
	    \WHILE{stopping criterion not met}
	        \STATE $X \gets a(\Theta, \widehat{g})$
	        \STATE $Y \gets \mathit{intensification}(X, g)$
	        \STATE $\mathit{update\_incumbent}(\widehat{\vec{\theta}}^*, X, Y)$
	        \STATE Update surrogate model $\widehat{g}$ with $(X,Y)$
	    \ENDWHILE
	    \RETURN $\widehat{\vec{\theta}}^*$
	\end{algorithmic}
\end{algorithm}

\paragraph{Surrogate models}
The task of the surrogate model, sometimes also called response surface or empirical performance model, is to approximate the original though unknown function $g$. Several classes of surrogate models have been considered in the literature, most of which are of a probabilistic nature in order to properly capture uncertainty. The most common ones are Gaussian processes (GP) \citep{williams2006gaussian}, tree Parzen Estimators \citep{bergstraBBK11} or random forests \citep{breiman01}. Due to their computational complexity, GPs are often rather impractical for AC problems when considering more than a handful of parameters to optimize.

\paragraph{Acquisition functions}
The acquisition function handles the exploration-exploitation dilemma common to optimization techniques. The function must decide to either return configurations from regions of $g$ that are likely to yield good performance according to $\widehat{g}$ (exploitation), or to return configurations from regions with larger uncertainty (exploration). A common criterion powering the acquisition function is expected improvement \citep{mockus74on,jones1998efficient} (EI), which in its simplest form\footnote{Note that for simplicity we assume that $g$ is deterministic here.} can be defined as 
\begin{equation}
    \mathbb{E}[I(\vec{\theta})] = \mathbb{E}\left[ \max\left\{0, g(\widehat{\vec{\theta}}^*) - g(\vec{\theta}) \right\} \right],
\end{equation} where $\widehat{\vec{\theta}}^*$ is the currently best known configuration, i.e., the incumbent. The expectation is required because $g(\vec{\theta})$ is unknown at the time of computation and hence, it is a random variable. Accordingly, to compute the EI, a \textit{probabilistic} surrogate model is required. For a detailed overview of different acquisition functions, we refer to \citep{frazier2018a}. 

\paragraph{Non-general AC SMBO approaches}
Several SMBO approaches exist to perform a limited form of AC, i.e., on only a single instance. We include these methods due to their historical importance to the field of AC, as well as because they may inspire new general AC approaches. Most SMBO based AC approaches are based on the idea of sequential kriging meta-modelling \citep{huang2006global} (SKO) and sequential parameter optimization (SPO) \citep{bartz2005sequential}, both of which are based on efficient global optimization \citep{jones1998efficient}. While the latter is a classical approach to black box function optimization using BO, both SPO and SKO constitute extensions to noisy black box functions; an assumption that is much more realistic for AC. However, both of these approaches still have potential drawbacks. Some of these are fixed by SPO$^+$ \citep{hutter2009experimental}, which improves the intensification scheme, and time-bounded SPO (TB-SPO) \citep{hutter2010time}, which generalizes SPO$^+$ to work under (potentially tight) time constraints instead of considering the number of function evaluations as a stopping criterion.

\subsection{General Model-based AC Methods}

\paragraph{SMAC}\label{SMAC}
Sequential model-based optimization for algorithm configuration (SMAC) \citep{hutter2011sequential,lindauer2021smac3}  can be seen as one of the first fully-fledged model-based AC approaches, as it features solutions for many of the limitations of the previously discussed SMBO techniques. SMAC generalizes TB-SPO to perform configuration over multiple problem instances so that it can support categorical parameters and handle tight time constraints. 

To support multiple problem instances, SMAC adapts the intensification strategy of TB-SPO to iteratively evaluate configurations on randomly sampled combinations of seeds and problem instances. When doing so, it ensures that configurations are compared only based on a performance estimate computed on the same randomly sampled set of problem instances. Furthermore, SMAC's surrogate model can generalize across problem instances by incorporating problem instance features. To this end, a surrogate model is learned on the joint problem instance and configuration space to predict the performance of a given configuration on a given problem instance. 

As a means to deal with a mixture of categorical and numerical parameters, SMAC employs a random (regression) forest as a surrogate model, which can inherently handle both types of parameters. The probabilistic nature of the surrogate model required for most acquisition functions is preserved by computing the predicted mean performance and the variance of a configuration as the corresponding statistics on the predictions of the individual trees. Moreover, SMAC leverages an adapted acquisition function. In particular, it uses a special form of the EI criterion and solves the EI maximization problem by a multi-start local search procedure that is equipped with special neighborhoods to deal with categorical hyperparameters. In addition to the configurations obtained through the local search procedure, SMAC evaluates the EI on a set of randomly sampled configurations for the purpose of exploration.

Anastacio and Hoos (\citeyear{anastacio2020model}) propose SMAC+PS, which integrates the idea of probabilistic sampling known from irace into SMAC. This enhancement yields improvements over both SMAC and irace in many cases. In particular, Anastacio and Hoos (\citeyear{anastacio2020model}) account for the problem that many of the completely randomly sampled configurations by SMAC often exhibit rather bad performance and thus, their evaluation yields only limited information. To this end, the authors suggest to sample configurations according to a truncated normal distribution centered around the default configuration.

In \citep{lindauerH18warmstarting} the authors suggest two different strategies to \emph{warmstart} model-based AC approaches and apply their suggestions to SMAC, leading to significant speedups from days to hours of configuration time. The idea underlying warmstarting is to use the evaluations of configurations from previous runs, i.e., on different problem instance sets, to speed up the configuration process in new runs of the configurator on a new set of instances.

Distributed SMAC \citep{hutter2012parallel} (D-SMAC) is an extension of SMAC leveraging parallelization to speed up the configuration process. The main idea behind D-SMAC is to parallelize target algorithm runs onto available workers as much as possible. For this purpose, it maintains a queue of target algorithm configuration evaluations to be performed, which is refilled such that the amount $k$ of available workers can always be used to its maximum. The intensification strategy is thus adapted to only push a set of evaluations to be performed into the queue instead of actually performing the necessary evaluations. While the workers are performing evaluations, a master process keeps track of the queue, selects configurations to be evaluated and updates the surrogate model. Furthermore, the authors replace EI by a parametrized upper confidence bound (UCB) criterion \citep{jones01a}, where the parameter controls the degree of exploration. To allow for an efficient generation of solution candidates based on the acquisition function, the master divides the queue into blocks of $k$ slots, i.e., one for each worker, and generates the corresponding configuration to be evaluated based on $k$ differently parameterized instantiations of the UCB criterion.

\paragraph{GGA++}\label{GGA++}
\citet{ansotegui15gga++} adapt the model-free AC approach GGA to include a surrogate model. More precisely, the authors use a surrogate model to evaluate the quality of new configurations. They integrate this within a crossover operator and call it genetic engineering. Recall that GGA contains both a competitive and non-competitive population in which winning configurations from the races between members of the competitive population are recombined with individuals from the non-competitive population. To this end, the crossover operator generates individuals according to the parameter tree crossover of the original GGA method and evaluates them using the surrogate. Note that rather than predicting the solution quality or runtime directly, the surrogate predicts the rank the individual would have in a tournament. The individuals with the best ranks are accepted into the population of the next generation in the same way as in GGA.

While the GGA++ surrogate is based on a random forest model, it differs in a key way. The premise of a random forest is to equally approximate the underlying function over the complete input space. In the case of AC, this is undesirable as only the areas of the input space that correspond to high-quality configurations are of interest. Thus, the authors present specialized splitting criteria that focuses on only the best configurations to increase the focus in high-quality regions of the configuration space while sacrificing approximation quality in other regions.

\paragraph{Graphical models}\label{GM}
\citeauthor{birattari2011learning} propose an approach we refer to as model-based graphical methods (MBGM) \citep{birattari2011learning} leveraging Bayesian networks (BNs) as a probabilistic surrogate model. Each node in such a network has an associated prior probability distribution over the parameter values. Here, both the network structure and the initial probability distributions of the nodes are supposed to be provided in advance. The approach then works as follows. First, an unseen problem instance from the set of training instances is chosen. Second, new configurations are created by sampling a value for each parameter one after another according to the conditional probability distribution modeled by the respective part of the BN. The order in which parameters are sampled is imposed by any of the topological orderings of the BN such that the sampling adheres to parameter dependencies. Third, the sampled configurations are evaluated on the unseen problem instance and, based on the performance, a subset of them is chosen to be added to a training dataset. Fourth, the probability distributions of the nodes are updated according to the posterior distribution based on the data contained in the training dataset. This process is repeated until a stopping criterion is reached and the best seen configuration is returned. 

A key aspect of the approach is the update procedure for the probability distributions of the nodes. This must ensure that the distributions are shifted towards more promising regions to yield good performance. Correspondingly, \citet{birattari2011learning} show how this update can be performed efficiently for numerical and categorical parameters and even for networks for parameters of mixed type (categorical and numerical). 

\paragraph{BNT}\label{BNT}
\citeauthor{do2020automatic} propose a method called Bayesian network tuning (BNT) \citep{do2020automatic}, which follows the same idea of leveraging BN for algorithm configuration, but employs a population of configurations. BNT starts by generating and evaluating an initial population of configurations according to a given initial design, and then iterates over the following steps. First, a subset of the current population corresponding to the best performing configurations is selected. Second, on the basis of this subset, a BN is created leveraging special metrics quantifying parameter interdependence. In contrast to \citep{birattari2011learning}, BNT does not assume a fixed network structure to be given, but creates it itself in each iteration.  Third, new configurations are created by sampling from the BN as described earlier. Finally, the newly created configurations are evaluated and replace the worst configurations in the current population. This process is repeated until a stopping criterion is reached. 

\paragraph{REVAC}\label{REVAC}
Parameter relevance estimation and value calibration (REVAC)~\citep{nannen2006method, nannen2007efficient} is an offline configurator for continuous parameters. It estimates probability density functions over parameter values, where the distribution can be used to draw conclusions about parameter relevance and parameter ranges after termination. It does not incorporate a capping mechanism, and continuous as well as categorical values need to be discretized. 

REVAC is based on the ideas of estimation of distribution algorithms and genetic algorithms. While creating and evaluating configurations, REVAC maintains a probability density distribution for each parameter. More precisely, REVAC starts with a uniform distribution, which is iteratively refined as configurations are created and evaluated, giving higher probabilities to parameter regions that perform well. New configurations are created by sampling from the estimated distribution, where a population approach is used that smoothens the distribution. The state of the probability model after termination is used to estimate the importance of parameters, and the Shannon entropy can be used to estimate the number of evaluations necessary to reach a defined target cost. Smit and Eiben (\citeyear{smit2009comparing}) add racing and a mechanism called sharpening to REVAC that, similar to GGA, gradually increases the number of instances a configuration is tested on. Another approach that utilizes evolutionary operators is given by~\citet{oltean2005evolving}, in which genetic programming is used on problem dependent C-programs to find configurations. The main drawback here is that crossover and mutation are dependent on the problem instance type (e.g., SAT, MILP, etc.).

\section{Theoretical Guarantees}\label{sec:Guarantees}

The field of AC began with heuristics that offered no guarantees as to the quality of the configurations found. Recently, there has been significant progress in providing a theoretical foundation to AC, offering bounds on the quality of the configurations found. First, we consider current results regarding the number of training instances and the structure of the class of cost functions guaranteeing good generalization of algorithm configurators. Then, we discuss and highlight the recent contributions focusing on the theoretical analysis with respect to the worst-case runtime of algorithm configurations.

\subsection{Generalization Guarantees} \label{subsec:gen_guarantees}
Most of the algorithms configurators encountered so far use the empirical mean as the aggregation function $m$ in \eqref{eq:2}, which in light of the objective function given by the expected costs in \eqref{eq:1} is a natural choice. This choice of aggregation function raises two urgent questions:
\begin{itemize}
	\item [1.]  Given some finite overall computational budget for each algorithm configuration at hand (i.e., a limit on the number of times a configuration can be run in total), how should this budget be distributed across the set of training instances to obtain the most accurate estimate via the empirical mean for \eqref{eq:1} for some configuration $\theta?$
	\item [2.] How many training instances are needed so that an arbitrarily small estimation error of the minimum \eqref{eq:1} can be guaranteed with high probability?
\end{itemize}

\paragraph{Distributing the computational budget}
\citet{birattari2004estimation} analyzes the first question under the assumption of an infinite set of training instances and a computational budget $N$ for a configuration $\theta.$ It is shown that one single run on $N$ many problem instances leads to the empirical mean estimate with minimal variance. However, the assumption of an infinite set of training instances is obviously quite restrictive for practical applications.

The latter result of \citet{birattari2004estimation} has been generalized by \citet{liu2020performance} to the more relevant scenario, in which the number of training instances is finite. They show that the empirical mean estimate with the minimal variance is obtained by distributing the available computational budget $N$ of a configuration as evenly as possible across the training instances. More precisely, if $K$ is the number of training instances, then the number of runs $n_i$ on a problem instance $i$ should be such that $n_i \in \{  \lfloor N/K \rfloor, \lceil N/K \rceil \}.$ Note that this distribution scheme is employed by some of the algorithm configurators discussed before, such as ParamILS, irace, and SMAC.

\paragraph{Probably approximately correct (PAC) learning}
The estimation error of the empirical mean estimate is usually set to be its absolute deviation from its population counterpart, i.e., for any $\vec{\theta} \in \Theta$ it is given by
\begin{align} \label{def:est_error}
	\Big|  \frac{1}{|\mathcal{I}_\mathit{train}|} \sum\nolimits_{i \in  \mathcal{I}_\mathit{train} } c(i,\vec{\theta}) - \int\nolimits_{\mathcal{I}} c(i,\vec{\theta})\,d\Pd(i)  \Big|.
\end{align}
With this in mind, the general approach to answer the second question is to derive high probability bounds on the estimation error \eqref{def:est_error} holding for any configuration $\vec{\theta} \in \Theta,$ which are monotonically decreasing with the number of training instances, i.e., $|\mathcal{I}_\mathit{train}|.$ Given a desired value of the estimation error, say $\varepsilon>0,$ one can obtain an answer to the second question by equating the derived bound and the estimation error, and then solve this equation with respect to the number of training instances.

Needless to say, these bounds will highly depend on the desired high probability level, the class of cost functions $\mathcal C_\Theta = \{i \mapsto c(i,\vec{\theta})  \, | \, \vec{\theta} \in \Theta \}$ or its dual function class $\mathcal{C}_\Theta^* =  \{\vec{\theta} \mapsto c(i,\vec{\theta})  \, | \, i \in \mathcal{I} \}$, as well as on the distribution of the problem instances $\Pd$ or the set of training instances $\mathcal{I}_\mathit{train}.$ If the latter dependency is taken into account the high probability bounds or equivalently the resulting guarantees for the number of training instances are said to be \emph{data-dependent}, and otherwise \emph{worst case} (aka \emph{distribution-free}) high probability bounds/guarantees. 

\cite{liu2020performance} derive worst case high probability bounds on the estimation error \eqref{def:est_error} for any $\vec{\theta} \in \Theta,$ if the configuration space $\Theta$ is finite and the cost function takes values only in some compact interval. Assuming that the functions in the dual function class $\mathcal{C}_\Theta^*$ are all Lipschitz continuous, the authors show high probability bounds on the estimation error if the configuration space is infinite, but bounded.

Guided by the observation that the shape of the cost functions in $\mathcal{C}_\Theta$ has a piecewise structure in many domains, i.e., piecewise-constant/linear, etc., \citet{balcan2021much} derive worst case high probability bounds on the estimation error \eqref{def:est_error} for any $\vec{\theta} \in \Theta$ for classes of cost functions exhibiting this structure. Such piecewise structures of the cost function have been observed for branch-and-bound AC problems \citep{balcan2017learning} or linkage-based hierarchical clustering algorithms \citep{balcan2018learning,balcan2020learninglink}. In a subsequent work, \citet{balcan2020refined} generalize their results by replacing the piecewise-structural assumption on $\mathcal{C}_\Theta$ by an $L^{\infty}-$norm approximation assumption of $\mathcal{C}_\Theta^*$. More precisely, it is assumed that any function in $\mathcal{C}_\Theta^*$ can be approximated uniformly over the configuration space $\Theta$ by a function from a class of functions having small Rademacher complexity\footnote{Rademacher complexity \citep{Bartlett2002} is a commonly used quantity in statistics and machine learning to measure the complexity of a function class.}. Unlike the previous worst case high probability bounds, the authors derive data-dependent high probability bounds based on the empirical Rademacher complexity of the class of functions approximating $\mathcal{C}_\Theta$ with respect to the $L^{\infty}-$norm. Roughly speaking, the result takes advantage of the fact that the Rademacher complexity of $\mathcal{C}$ is small if the Rademacher complexity of the class of functions approximating its dual function class $\mathcal{C}_\Theta$ is small. However, it is shown that the latter fact relies heavily on the approximation with respect to the $L^{\infty}-$norm, as it is also shown that it is impossible to derive non-trivial generalization guarantees if the approximation holds only under the $L^{p}-$norm with $p<\infty$. This is due to the fact that a small Rademacher complexity of the class of functions approximating $\mathcal{C}_\Theta$  with respect to the $L^{p}-$norm with $p<\infty$ does not imply a small Rademacher complexity of $\mathcal{C}.$

\subsection{Runtime Analysis}

Initiated by the work of \citet{kleinberg2017efficiency}, the AC problem has recently attracted a great deal of research focusing on theoretically grounded approaches regarding the design and motivation of a configurator, if the relevant cost function $c$ (see Section \ref{Problemsetting}) considered is the runtime of configuration $\vec{\theta}$ on problem instance $i$. These approaches take into account, on the one hand, how close the supposed best algorithm configuration returned by the configurator is to the actual optimal configuration and, on the other hand, how long it takes on average to return it in the worst case. For this, a definition of an optimal algorithm configuration is required, as is a measure of closeness. Assuming that problem instances are drawn with some distribution $\Pd$ from $\I,$ the expected runtime of a configuration $\vec{\theta}$ is $R(\vec{\theta})= \mathbb{E}_{i \sim \Pd }(c(i,\vec{\theta})).$ Thus, the configuration with the optimal expected runtime is given by $ \argmin_{\vec{\theta}  \in \Theta } R(\vec{\theta})$ having (optimal) expected runtime 
\begin{equation*}
    \mathrm{OPT} := \inf_{\vec{\theta}  \in \Theta } R(\vec{\theta}).
\end{equation*}
The search for the optimal configuration is generally too ambitious, as the total runtime required for the configurator must be extraordinarily large (possibly infinite) to guarantee that the best algorithm configuration returned by the configurator is in fact the optimal one with high probability.

As a workaround, one can leverage the idea underlying PAC learning \citep{valiant1984theory} to the problem at hand. The basic idea is to relax the goal of finding the optimal configuration itself and, instead, find a configuration that is considered to be ``good enough''. As there are potentially several such ``good enough'' configurations\footnote{Of course, the optimal configuration itself is always ``good enough''.}, this relaxation of the goal allows the search to be completed in less (and, thus, feasible) time. In this context, ``good enough'' means that the expected runtime is only worse than the optimal expected runtime up to a multiplicative factor of $1+\varepsilon$ for some fixed precision parameter $\varepsilon>0.$ Formally, a configuration is said to be $\varepsilon$-optimal (``good enough'') iff 
\begin{equation*}
    \mathbb{E}_{i \sim \Pd }(c(i,\vec{\theta}) ) \leq (1+\varepsilon)  \mathrm{OPT}.
\end{equation*}
However, this relaxation of the target is problematic in the context of AC problems, since the runtimes of configurations often exhibit a heavy-tailed distribution. Indeed, it is not difficult to construct an example based on such distributions in which any (sensible) configurator would, in the worst case, take infinitely long to find an $\varepsilon$-optimal configuration; see for instance \cite[Example 11.1.1]{vitercik2021automated}.

In light of the prevalence of heavy-tailed distributions in the realm of AC problems, the remedy is to run configurations only up to some timeout $\kappa\geq 0$ chosen by the configurator that could vary over the runs. In practice, this means that one observes $\min\left(c(i,\vec{\theta}), \kappa\right)$ if configuration $\vec{\theta}$ is run on $i$ with timeout $\kappa,$ and also whether the configuration runs into a timeout or solves the problem instance within $\kappa.$ This gives rise to the $\kappa$-capped expected runtimes of a configuration $\vec{\theta}$ defined by $R_\kappa(\vec{\theta}) = \mathbb{E}_{i \sim \Pd }(\min\left(c(i,\vec{\theta}), \kappa\right)),$ which now take the place of the uncapped runtimes $R(\vec{\theta})$\footnote{Technically, $R(\vec{\theta}) = R_\infty(\vec{\theta}) $.}. However, the introduction of timeouts inevitably means that a certain proportion of problem instances may not be solved by a configuration within the given timeout. This immediately raises the question of how to choose the timeouts so that the proportion of unresolved problem instances is tolerable, but at the same time, the ($\kappa$-capped) expected runtime is not too far from the optimal one? The notion of $(\varepsilon,\delta)$-optimality of a configuration is introduced by \citet{kleinberg2017efficiency} and provides an intuitive answer. A configuration is $(\varepsilon,\delta)$-optimal if there is a timeout $\kappa\geq 0$ at least as large as the $\delta$-quantile of the configuration's runtime distribution, such that the configuration's $\kappa$-capped expected runtime is at most $(1+\varepsilon)\mathrm{OPT}.$ 
Formally, 
\begin{align} \label{defi:eps_delta_optim}
	 \mbox{$\vec{\theta}$ is $(\varepsilon,\delta)$-optimal } \ \Leftrightarrow \ \exists \kappa\geq 0: 	R_\kappa(\vec{\theta})\leq 	(1+\varepsilon)  \mathrm{OPT}  \wedge  \mathbb{P}_{i \sim \Pd} (c(i,\vec{\theta}) >\kappa) \leq \delta.	
\end{align}
This notion of $(\varepsilon,\delta)$-optimality has been adopted by subsequent works and slightly modified in some cases. Nonetheless, the core idea remains the same, namely that one is willing to leave a fixed share of problem instances unsolved (captured via the $\delta$-quantile) to improve the expected runtime, such that it is close to optimal (captured by the additive term $\varepsilon  \mathrm{OPT}$). \citet{kleinberg2017efficiency} show that the worst-case expected runtime of any configurator to return an $(\varepsilon,\delta)$-optimal configuration (with probability of at least $1/2$) is of order $\Omega\big( \frac{|\Theta|}{\delta \varepsilon^2} \mathrm{OPT} \big),$ when the number of configurations is finite.

\paragraph{From finite to infinite number of configurations.}
To obtain a lower bound result for the case of a large or even possibly (uncountable) infinite number of configurations, the notion of $(\varepsilon,\delta)$-optimality needs to be modified, as it is the main limiting factor in this regard. Indeed, due to its definition in \eqref{defi:eps_delta_optim}, any optimal configurator is forced to explicitly consider each configuration to decide on its $(\varepsilon,\delta)$-(sub-)optimality. To this end, in a scenario with a large number of configurations, the goal is relaxed to find a $(\varepsilon,\delta)$-optimal configuration after excluding    the $\gamma \in(0,1)$ fraction of best configurations from $\Theta$ with respect to the expected runtime. Formally, let $\mathrm{OPT}_{\gamma} = \inf_{x \in \mathbb{R}_+} \{ x \, | \, \mathbb{P}_{\vec{\theta} \sim \mathrm{Unif}(\Theta)} (R(\vec{\theta}) \leq x) \geq \gamma  \}$ be the optimal expected runtime after excluding the $\gamma \in(0,1)$ fraction of best configurations, then a configuration $\vec{\theta}$ is $(\varepsilon,\delta,\gamma)$-optimal iff
\begin{align} \label{defi:eps_delta_gamma_optim}
	\exists \kappa\geq 0: 	R_\kappa(\vec{\theta})\leq 	(1+\varepsilon)  \mathrm{OPT}_{\gamma}  \wedge  \mathbb{P}_{i \sim \Pd} (c(i,\vec{\theta}) >\kappa) \leq \delta.	
\end{align}
\citet{kleinberg2017efficiency} provide a result on the worst-case expected runtime of any configurator to return an $(\varepsilon,\delta,\gamma)$-optimal configuration for specific choices of $\delta$ and $\gamma,$ which is quite similar to the finite case by replacing $\mathrm{OPT}$ by $\mathrm{OPT}_\gamma$ and $|\Theta|/\delta$ by a parameter common to the choices of $\delta$ and~$\gamma.$

Another convenient tool provided by the authors is how to turn a configurator with theoretical guarantees for finding $(\varepsilon,\delta)$-optimal configurations for a finite configuration space to finding $(\varepsilon,\delta,\gamma)$-optimal configurations for an infinite configuration space. This can be done via uniform sampling from $\Theta$ as follows. If a configuration is sampled uniformly at random from $\Theta,$ then obviously this configuration belongs with probability $\gamma$ to the $\gamma$ proportion of the best configurations, i.e., the best $\gamma$ configurations. Thus, if $n$ many configurations are sampled uniformly at random from $\Theta,$ then the probability that at least one among these $n$ many belongs to the best $\gamma$ configurations is at least $1-(1-\gamma)^n.$ Now, fixing a failure probability $\zeta \in(0,1)$ for the non-occurrence of the latter event, one can solve the resulting inequality with respect to $n$ to obtain that $ \lceil \frac{\log(\zeta)}{\log(1-\gamma)} \rceil $ samples are sufficient to guarantee that at least one configuration within the sample belongs to the best $\gamma$ configurations with probability at least $1-\zeta.$

\citet{balcan2020learning} provide an alternative way to the uniform sampling approach to obtain a finite set, which includes at least one configuration with ``good enough'' performance. 
Here, ``good enough'' performance is again in terms of $(\varepsilon,\delta)$-optimality of a configuration, which, however, is defined in a slightly different way as in \eqref{defi:eps_delta_optim}, namely
\begin{align} \label{defi:eps_delta_optim_mod}
	\mbox{$\vec{\theta}$ is $(\varepsilon,\delta)$-optimal } \ \Leftrightarrow \ 	R_{t_{\vec{\theta}}(\delta)}(\vec{\theta})\leq 	(1+\varepsilon) \mathrm{OPT}_{\alpha \delta},	
\end{align}
where $\mathrm{OPT}_{\alpha \delta} = \min_{\vec{\theta}} R_{t_{\vec{\theta}}(\alpha \delta)}(\vec{\theta}),$ $t_{\vec{\theta}}(x)$ is the $x$-quantile of $\vec{\theta}$'s runtime distribution and $\alpha\in(0,1)$ is a slack parameter that is introduced for details outside the scope of this work\footnote{This alternative notion of $(\varepsilon,\delta)$-optimality is due to \cite{weisz2019capsandruns}.}. It is worth noting that the authors allow other cost functions than runtime in their work, as the notion of $(\varepsilon,\delta)$-optimality can also be used for such variants. Moreover, the theoretical guarantees as well as the design of their approach assume that the cost functions $\mathcal{C}_\Theta$ (cf.\ Subsection \ref{subsec:gen_guarantees}) are piecewise constant.

\paragraph{Structured procrastination (SP)}\label{Structured}
\citet{kleinberg2017efficiency} propose an AC technique in which the idea is to postpone potentially hard problem instances and solve supposedly easier problem instances first. It thus only spends time on hard instances if it is unable to solve easy ones, reducing the overall time the technique needs. The authors prove that SP returns an $(\varepsilon,\delta)$-optimal configuration with high probability, and that the runtime is optimal up to a logarithmic factor.

We briefly describe how SP works for large parameter spaces, as this is the more realistic case of the approach for practical applications. The approach uses a double-ended queue $Q_\theta$ that stores (instance, timeout) pairs for each configuration. A pair is pulled from the front of the queue and run. Should the configuration not finish the instance within the timeout, the instance is placed at the back of the configuration's queue with double the timeout. If the instance is completed within the timeout, this information is saved and the instance is not considered again for configuration $\theta$. In each iteration of the approach, the $Q_\theta$ is chosen with a $\theta$ with minimal average performance observed so far. To avoid requiring the user to specify $\delta$, SP is implemented as an anytime algorithm that reduces $\delta$ over the course of its execution. Note that SP returns the configuration with the longest total execution time rather than the best empirical mean due to theoretical reasons.

\paragraph{LeapsAndBounds}\label{LeapsAndBounds}
\citet{weisz2018leapsandbounds} propose a phase-based algorithm configurator called LeapsAndBounds that tries to guess an appropriate upper bound on the optimal runtime in each phase by doubling the guess of the bound after each failed phase. The authors provide an upper bound on the worst-case total amount of time their method needs for finding an  $(\varepsilon,\delta)$-optimal configuration with high probability, which improves upon the result of SP. Furthermore, the superiority of LeapsAndBounds over SP is confirmed empirically on a benchmark of SAT solvers.

In each phase of LeapsAndBounds, each configuration is given a time budget, and a number of problem instances to be solved within the budget. Instances are chosen depending on the phases that have already passed, with specific choices of the phase-dependent quantities being based on empirical Bernstein stopping \citep{mnih2008empirical}. This mechanism takes the range and the empirical variance of capped runtime observations into account.
If a configuration exhausts the time budget without having solved all problem instances, its expected runtime is above the phase-dependent upper bound guess with high probability. If this happens for all configurations, the phase has failed and the next phase is started. In the case that some configurations do not manage to use the time budget completely, the empirical mean of the runtimes is accepted as a suitable estimator for the expected runtime, and the configuration with the lowest mean is returned as an $(\varepsilon,\delta)$-optimal configuration. Note that even though the time budget is not exhausted, there may still be instances that hit their timeouts.

\paragraph{Structured procrastination with confidence}\label{SPwC}
\citet{kleinberg2019procrastinating} revises the SP configurator and modifies the selection and return criterion used to choose the configuration to recommend at each step to improve its practical performance while maintaining satisfactory theoretical guarantees. More precisely, the authors derive lower confidence bounds on the expected mean runtime of a configuration, which then take over the role of the empirical mean runtimes in the selection process\footnote{This modified selection rule adopts the \emph{optimism in the face of uncertainty principle}, which is a popular paradigm in the realm of reinforcement and bandit learning problems \citep{lattimore2020bandit}}. Thanks to a careful derivation of these lower confidence bounds, the choice mechanism adapts to the difficulty of the configuration problem such that poorly performing configurations are excluded quite quickly; a crucial property that SP, in general, lacks. In addition, the lower confidence bounds, as well as the minimal length of a configuration's queue, are chosen such that $\epsilon$ and $\delta$ do not need to be specified a priori, which, however, is required by all methods discussed before. Another change to SP is that the configuration returned is the one having the highest number of completed as well as pending tasks in its queue. The rationale behind this modification is that more promising configurations run faster on average and thus have a larger number of completed and pending tasks. The resulting modified version of SP is called SP with confidence (SPC) due to the usage of the (lower) confidence bounds.

The authors show the correctness of SPC and derive an improved runtime bound over SP. In addition, they show in an experimental study for a simple benchmark set of SAT solvers that SPC finds reasonable configurations after a smaller amount of computation time than SP and LeapsAndBounds.

\paragraph{CapsAndRuns}\label{CapsAndRuns}
As a follow-up to LeapsAndBounds, \citep{weisz2019capsandruns} propose CapsAndRuns, which improves the former both theoretically and practically. It is also based on Bernstein stopping~\citep{mnih2008empirical}, but extends LeapsAndBounds in how it estimates timeouts. Furthermore, CapsAndRuns refines the upper bound on the total time needed for configuration and adjusts the notion of $(\varepsilon,\delta)$-optimality as specified in \eqref{defi:eps_delta_optim_mod}. More precisely, CapsAndRuns first estimates a timeout for each configuration and performs a Bernstein race over the configurations afterward.%

CapsAndRuns works in two phases, but note that these phases are different from the phases in LeapsAndBounds. First, the method estimates a timeout for each configuration such that only a $\delta$-quantile of the instances exceed the timeout, followed by a phase using this estimate to return an expected runtime estimate using the estimated timeout. To make the search process even more efficient, a global estimator for the optimal expected runtime (considering timeouts) is used for all configurations across both phases to eliminate suboptimal configurations as early as possible. This elimination happens either directly after the first phase, when it turns out that the configuration is too slow, or during the second phase, as soon as one is confident enough that the configuration's empirical mean runtime is above the optimal one. A further aspect of the approach is that it can be parallelized across configurations.

In other words, the (sub-)optimality of configurations is measured by means of the closeness of their expected runtimes capped at their $\delta$-quantile to the optimal expected  $\alpha \delta$-quantile-capped runtime.

\paragraph{ImpatientCapsAndRuns}\label{ImpatientCapsAndRuns}
Guided by the observation that heuristic configuration approaches achieve appealing practical performance by quickly discarding less promising configurations based on only a few observations of their runtimes, \citet{weisz2020impatientcapsandruns} propose the ImpatientCapsAndRuns (ICAR) algorithm, which builds on CapAndRuns with a more aggressive elimination strategy. This is achieved through a preprocessing mechanism for filtering configurations that are unlikely to be optimal that is run before entering CapsAndRuns. \citet{weisz2020impatientcapsandruns} prove that ICAR finds an $(\varepsilon,\delta,\gamma)$-optimal configuration with high probability for specific ranges of $\varepsilon,\delta$ and $\gamma$.

The preprocessing technique in ICAR is essentially a stripped down version of the two phases of CAR, except that the internal statistics for elimination are chosen differently. Note that this still depends on the global estimate of the optimal expected runtime (with timeouts). For this reason, the preprocessing and the subsequent two phases of CAR are executed successively one after the other on an ever-increasing pool of configurations (batches), with configurations that have already been eliminated no longer being taken into account. This ensures that the global estimator gradually improves, which, in turn, leads to the pre-elimination by the preprocessing routine becoming more and more precise and aggressive. 

As a byproduct of the theoretical analysis of ICAR, the authors improve the CAR algorithm by refining the choice of one of its key internal statistics. Furthermore, experimental studies show that ICAR has significantly better practical performance for three benchmark datasets compared to the original CAR and its improved version.

\paragraph{ParamRLS with RLS$_k$} \label{ParamRLS}
Since the ParamILS algorithm (see Section \ref{sec:ModelFree}) does not provide any theoretical grounding, \citet{Hall2019Cutoff} introduce ParamRLS, which is a simplified version of ParamILS. The main difference is that ParamRLS uses a random local search excluding restarts instead of an iterated local search. This modification allows the authors to analyze theoretically the impact of the timeout on the expected number of required configuration evaluations. They prove that at least a timeout of $\Omega(n \log n)$ is necessary for a problem of size $n$ while tuning the target algorithm RLS$_k$ considering the configuration time for the OneMax problem function class. Moreover, they show that at least a timeout of $\Omega(n^2)$ is needed for the Ridge$^*$ problem function class. In addition, they identify $k=1$ as optimal for the OneMax function class when optimizing the fitness values of the configurations.

Specifically, ParamRLS initializes the configurator randomly and then increases or decreases a single parameter chosen uniformly at random by $1$ or by $2$, also uniformly at random, in each time step. This step is repeated until no parameter change yields an improvement anymore. Two different variants are considered. First, ParamRLS-T, in which the target metric is the optimization time, and second, ParamRLS-F, which identifies the configuration that yields the best-found fitness value. The local search algorithm RLS$_k$ has only one parameter $k$, which gives the number of bits in the configuration that are flipped in each iteration of the search.

\paragraph{ParamRLS with $(1+1)$EA}
After the successful theoretical analysis of the simple random local search, \citet{Hall2020} consider a more complex AC scenario by tuning the mutation rate $\chi$ of the target algorithm $(1+1)_{\chi}$EA with ParamRLS. $(1+1)_{\chi}$EA works by flipping each of the $n$ bits of the current configuration independently with probability $\chi/n$. Once again, the authors analyze the required timeouts for the runtime metric and the best found fitness value on the two benchmark problem classes Ridge and LeadingOnes. They further prove that all configurators which are using the runtime as a performance metric require a timeout at least as large as the expected time to identify the optimal configuration. Thus, this problem scenario needs larger timeouts than in the case of the best fitness performance metric.

\paragraph{Harmonic mutation operator}
Inspired by the insights from the above analyses of ParamRLS, \citet{Hall2020HM} design a harmonic mutation operator for the configurations that provably leads to faster performance in the case of single parameter target algorithms. In fact, the authors prove that it tunes single-parameter algorithms in polylogarithmic time for (approximately) unimodal parameter spaces. Even in the worst case, the harmonic mutation operator only slows down the algorithm by at most a logarithmic factor. In an experimental analysis, the harmonic mutation operator is shown to be superior to the $l$-step and random mutation operators used for ParamRLS and ParamILS.

To be more precise, the harmonic mutation operator selects a parameter uniformly at random and samples a step size according to the harmonic distribution. In fact, the probability for a step size $d$ is $1/(d\times H_{\Phi-1})$ with $H_m$ as the $m$-th harmonic number $H_m = \sum_{k=1}^m \frac{1}{k}$ and $\Phi$ as the range of possible parameter values. Then the best parameter value at distance $\pm d$ is returned. 

\section{Realtime Methods} \label{sec:Realtime}
Realtime methods relax the assumption made by offline methods that a representative training set is available since, in reality, such training sets might not always be available or costly to obtain. Due to possibly changing business models and requirements, they further allow $\Pd_t$, and therefore problem instances that are drawn from it, to change over time. This phenomenon is equivalent to concept drift in machine learning~\citep{gama2014survey}. Ultimately, this may lead to a growing disparity between $\hat{\vec{\theta}}$ and $\vec{\theta}^*$, possibly resulting in diminishing performance of the offline tuned configuration in production. Realtime algorithm configurators have been developed that can provide configurations on an ongoing, per-instance basis. That is, the training set $\mathcal{I}_\mathit{train}$ is not needed upfront, but problem instances are solved sequentially as they arrive and are used for configuration adjustments on the fly. In terms of their design, realtime configurators incorporate the same components discussed for offline configurators.

For the sake of completeness, we formulate the realtime configuration problem, which sometimes is also referred to as online tuning or parameter control, more formally. We are interested in a configurator $o : \mathcal{H} \times \I \rightarrow \Theta$ that for a given time step $t$ and the corresponding problem instance $i_t \sim \Pd_t$ provides a configuration $\vec{\theta}$ based on the history $h_t$. The history $h_t$ consists of triplets of the form $\{(i_k, \vec{\theta}_k, c_k)\}_{k=1}^{t-1}$ containing the problem instances encountered so far, the configurations used and the resulting cost. The goal is to minimize the (average) cost, which for a final time horizon is given by: $\mathcal{M}= \frac{1}{T} \sum_{t=1}^{T} c(i_t,\vec{\theta}_t)$. For a new problem instance, the optimal configuration in light of the instances encountered so far is then defined as
\begin{equation}
    \hat{\vec{\theta}}^{*} \in \arg\min_{\vec{\theta} \in \Theta} \E [c(i_t,\vec{\theta})| h_t] \,\,\, .
\end{equation}

\begin{figure}[ht]
  \centering
  \includegraphics[width=0.7\textwidth]{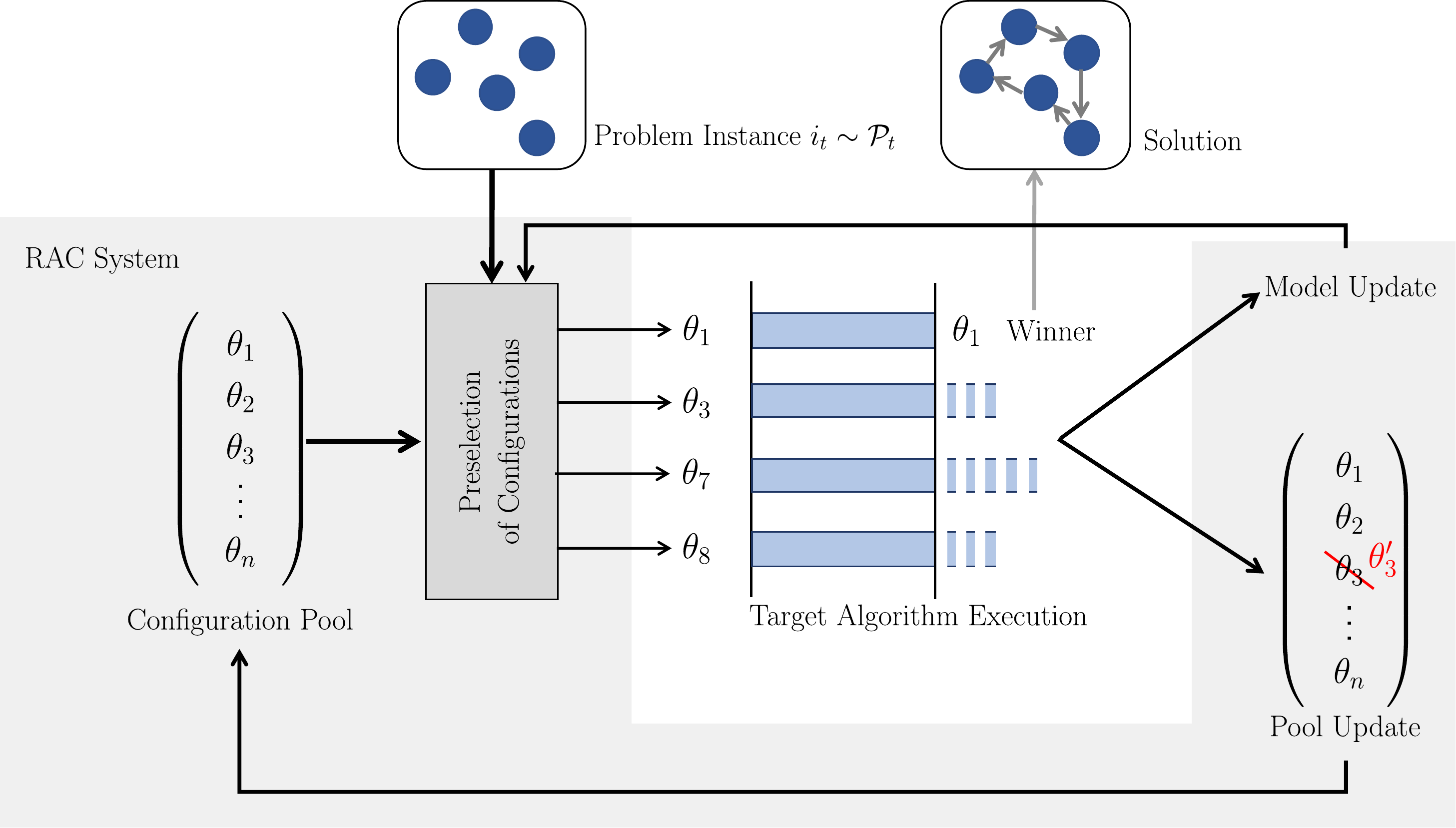}
  \caption{\label{fig:realtime} In realtime AC instances arrive and are solved sequentially}
\end{figure}

\paragraph{ReACT and ReACTR}\label{ReACT}
The first work to tackle the realtime configuration problem is the method realtime algorithm configuration through tournaments (ReACT)~\citep{fitzgerald2014react}. ReACT is a model-free configurator that leverages tournaments of configurations run in parallel for each incoming instance. It maintains a population of configurations and adjusts this in each iteration. It can handle continuous, discrete, and categorical variables. The extension ReACTR~\citep{fitzgerald2015reactr} introduces a ranking mechanism based on TrueSkill~\citep{herbrich2006trueskill} to rank configurations. This allows for a population that is larger than the number of CPU cores available, leading to more diversity. 

While ReACT and ReACTR share tournaments, they differ in population size, configuration elimination and generation mechanisms. Similar to GGA~\citep{ansotegui2009gender} both ReACT and ReACTR use parallel tournaments to evaluate individuals and assume that the configuration objective is runtime minimization. Thus, as soon as one of the target algorithm runs finishes, the finisher is declared the winner and all other runs are terminated. While ReACT only allows for a population size equal to the number of CPU cores, ReACTR allows for a bigger population. To avoid having to run multiple tournaments like GGA, ReACTR only runs the top-ranked configurations on the available cores, limiting the exploration of different configurations. Based on the information obtained through the tournament, weak configurations are removed from the pool. ReACT removes configurations based on domination, where it is ensured that a configuration was given enough opportunity to prove its worthiness by requiring another configuration to beat it at least $m$ times. Individuals are then replenished by random sampling. ReACTR, in contrast, uses its TrueSkill ranking mechanism to determine which individuals to remove. In addition, it supplements the random sampling of ReACT for making new configurations with a crossover and mutation procedure between the top-ranked individuals. Thus, it can more effectively intensify its search for good configurations around those that have worked well in the past, similarly to I/F-Race or GGA.

\paragraph{CPPL}\label{CPPL}
The contextual preselection with Plackett-Luce (CPPL) algorithm introduced by \citet{el2020pool} uses preselection bandits~\citep{bengs20a} to rank, choose and generate configurations while building on the racing and parallel execution principles from ReACTR~\citep{fitzgerald2015reactr}. It is a model-based, instance-specific configurator that works under the assumption of a Plackett-Luce model~\citep{plackett1975analysis,luce2012individual}.
Configurations are interpreted as slot machines (``bandits'') that have arms that can be ``pulled'' by running a configuration to observe their quality. Unlike the classical multi-armed bandit setting \citep{lattimore2020bandit}, where one arm is pulled resulting in a numerical observation (i.e., what is the quality of the pulled arm?), in the preselection bandit setting it is allowed to pull more than one arm at a time leading to qualitative feedback, such as winner feedback (i.e., which of the pulled arms had the highest quality?) or (partial) ranking feedback. Thus, it is an extension of the dueling bandit setting \citep{bengs2021preference} where only two arms are pulled (or dueled against each other).
The contextual preselection bandit extension~\citep{mesaoudi2020online} allows CPPL to take problem instance features as additional information into account. That is, the features of a problem instance provide CPPL with information on which arm (configuration) possibly performs best and therefore should be ``pulled''.

CPPL builds on ReACTR, but uses the bandit model for key operations.  More precisely, the bandit model selects a set of configurations that are to be raced against each other, replacing the TrueSkill ranking mechanism of ReACTR. Based on the obtained (censored) winner feedback, the model is updated using stochastic gradient descent. While ReACTR prunes configurations from the pool using TrueSkill, CPPL uses the upper confidence bounds on the estimated performance of the configurations for pruning. New configurations are created by choosing top-ranked configurations from the pool and combining them by means of a genetic engineering procedure as in GGA++, just with a different surrogate and including mutation after the crossover procedure. 

\section{Instance-specific Methods}\label{sec:Instancespecific}
Offline configurators like ParamILS, irace, GGA, or SMAC employ a one-size-fits-all paradigm, and while this works well for homogeneous problem instance sets where a single configuration yields good performance for all problem instances, it may fall short when the instance set is heterogeneous. For such situations, instance-specific configurators have been developed that can provide a configuration $\hat{\vec{\theta}}_i$ for a specific problem instance $i \sim \Pd$.

\begin{figure}[tb]
  \centering
  \includegraphics[width=0.8\linewidth]{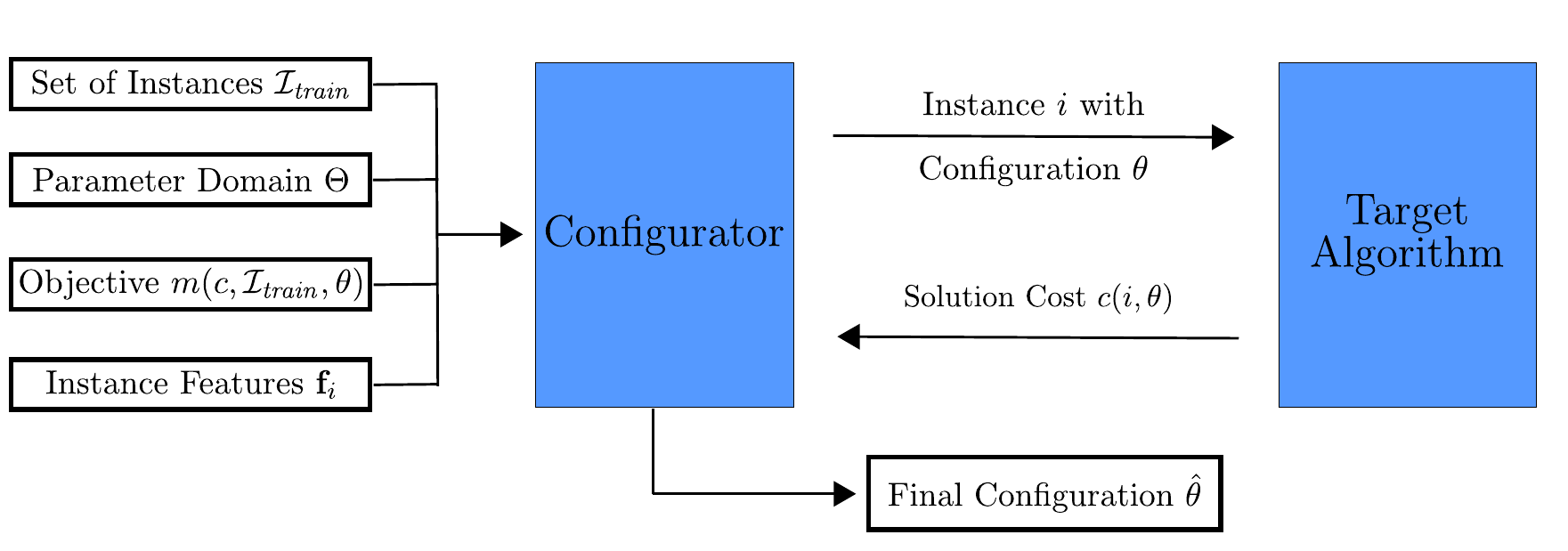}
  \caption{\label{fig:per_instance} Illustration of Instance Specific Configuration}
\end{figure}

Instance specific configurators require the same inputs as offline methods with the addition of a feature vector $\vec{f}_i$ for each problem instance (see Figure \ref{fig:per_instance}). By altering Equation~\eqref{eq:2}, the problem of instance specific configuration can formally be expressed as:
\begin{equation}
    \hat{\vec{\theta}}^{*} \in \arg\min_{\vec{\theta} \in \Theta} \E [c(i,\vec{\theta})] \,\,\, ,
\end{equation}
where $\E[c(i,\vec{\theta})]$ is the expected cost of $\A$ on a specific problem instance $i$ for which a configuration should be found. Configurators achieve this goal by harnessing the instance features and learning a relationship between the structure of the instances as described by the features and the performance of various configurations. Thus, the features must provide a good enough representation of the instances such that a model can be constructed, and they need to be correlated with the performance of the target algorithm. This is a difficult task that has itself been the focus of various work~\citep{kroer2011feature,tierneyalgorithm}. 

There are two main ways of generating instance-specific configurations:
\begin{inparaenum}[(1)]
\item incorporate the problem instance features into the existing prediction models of the configurator for configurations or 
\item deploy separate models that only consider the problem instance features without the configuration characteristics.
\end{inparaenum}
Both variants can be referred to as model-based, but this section focuses primarily on methods that do not utilize a surrogate model. In Section~\ref{sec:ModelBased} and~\ref{sec:Realtime} model-based and realtime configurators that are instance-specific and use configurations characteristics besides problem instance features in surrogate models are discussed.

\paragraph{ISAC}\label{ISAC}
Instance-specific algorithm configuration (ISAC)~\citep{kadioglu2010isac} and its extensions pair an algorithm configurator with a clustering algorithm. The clustering algorithm partitions the instance space, and the configurator (usually GGA/GGA++) is applied to each of the clusters to generate a configuration specific to the cluster. In the test phase, the learned configurations are assigned to the instances needing to be solved. The assignment mechanism varies depending on the version of ISAC that is used, and is described below in more detail.

ISAC uses the \emph{g-means}~\citep{hamerly2004learning} clustering algorithm with a minimum cluster size to generate clusters with the Euclidean distance as a metric. Note that $g$-means does not require the number of clusters to be specified in advance as in other clustering algorithms. Instead, it assumes that clusters ought to be normally distributed and splits clusters that are not using $k$-means with $k=2$. Once the clusters are determined, GGA~\citep{ansotegui2009gender} is used to find a representative configuration for each cluster identified by $g$-means, although any configurator could be used. In addition, an extra, default configuration is determined for all instances as a fallback mechanism for instances in the test phase that are not close to the previously determined clusters.  In the testing phase, when a new problem instance arrives, ISAC computes its proximity to each cluster based on the Euclidean distance and uses the configuration of the closest cluster or the default configuration to solve the problem instance. 

\citet{malitsky2014evolving} introduce evolving instance-specific algorithm configuration (EI-SAC), which is a retraining mechanism for the test phase that allows for reassignment of configurations to clusters as problem instances arrive. EISAC performs a re-clustering utilizing newly seen instances in addition to the previous training instances, where the update is performed based on the Rand index~\citep{rand1971objective} or when new configurations (solvers) are added/removed. Clusters are updated through solving an optimization problem that redistributes the configurations among clusters by minimizing the solving time of problem instances within a cluster. This means no new configurations are needed. However, additional target algorithm runs would still be needed for each cluster/configuration combination. To avoid this and to save further time, EISAC uses an empirical hardness model to predict the runtime of a configuration on new problem instances. Using this model makes the actual target algorithm runs only necessary when the problem instance would be assigned to a different cluster based on the cost prediction. 

Malitsky and Sellmann (\citeyear{malitsky2012instance}) extend ISAC to AS. ISAC is used to select an algorithm for a problem instance, and also configures the algorithm automatically, on a per cluster basis. A model-based algorithm selector is added by~\citet{ansotegui2016maxsat} in a later version, resulting in the ISAC++ method, which uses cost-sensitive hierarchical clustering (CSHC)~\citep{malitsky2013algorithm} as a selector.

\paragraph{\CluPaTra}\label{CluPaTra}
Another offline configurator that, similar to ISAC, derives configurations for clusters of problem instances is \CluPaTra~\citep{lau2011instance,lindawati2013clustering}. It is specifically designed for target algorithms that provide search trajectories. It utilizes AGNES~\citep{kaufman2009finding} for clustering and ParamILS as a configurator. Instead of using precomputed problem instance features like ISAC, \CluPaTra uses the search trajectories derived through runs of the target algorithm encoded as directed sequences. The search trajectory tracks the history of solutions found by the target algorithm through the search space, meaning \CluPaTra pierces the black box assumed by most other configurators. A sequence alignment method computes the similarity between trajectories for the clustering. A major drawback of \CluPaTra, and its extension, is that it is not fully clear how to utilize it in production. Determining the cluster membership of a new instance is a chicken-and-egg problem, as one must solve the instance to determine the search trajectory that is used for clustering.

The SufTra and FloTra methods extend \CluPaTra. The main focus for these approaches lies in different ways of integrating the search trajectories into the clustering. SufTra~\citep{yuan2013automated} replaces the directed sequence encoding and sequence alignment with a suffix tree encoding from which features are extracted through frequent substring alignment. The similarity between trajectories is computed using frequent substrings and the cosine similarity. In addition, easy problem instances that are similar to hard problem instances are used as surrogates during configuration by finding configurations on easy instances and using them on similar hard instances. This mechanism, however, again struggles from the previously mentioned chicken-and-egg problem, since the runtime needs to be known before being able to partition instances into easy and hard ones. In FloTra~\citep{lindawati2013flotra}, the use of the search trajectories is altered by creating graphs out of the trajectories, to which pattern mining is applied to derive features that are used for clustering.

\paragraph{Hydra}\label{Hydra}
\citet{xu2010hydra} and~\citet{xu2011hydra} introduce Hydra, which essentially adapts the boosting paradigm~\citep{schapire2003boosting} to algorithm configuration. Hydra iteratively configures a solver and adds the configuration to a portfolio that improves upon the weakness of the current portfolio. Hydra utilizes ParamILS~\citep{hutter2009paramils} to propose configurations to SATzilla~\citep{xu2008satzilla} which in turn constructs the portfolio. In general, any configurator and any AS technique can be used. The cost function of ParamILS scores configurations by their real value if they perform better than the current portfolio, and by the costs of the portfolio if not. This adjustment effectively penalizes configurations that yield a bad objective value on problem instances the portfolio also performs badly, while not penalizing bad objective values on problem instances on which the portfolio already performs well on. Hydra excels at specifically targeting areas of the feature space in need of custom configurations. However, this comes at a high computational cost as the configurations must be computed sequentially, unlike in ISAC, where they can be parallelized.

\paragraph{MATE}\label{MATE}
Genetic programming forms the basis for the model-based algorithm tuning engine (MATE) \citep{yafrani2020mate}. The approach takes problem instance features into account and provides human-understandable relations between features and target algorithm parameters. In particular, MATE uses symbolic regression~\citep{augusto2000symbolic} and tree-based genetic programming. Configurations are encoded as trees, which are compared to each other using a score function that is used to aggregate results over algorithm runs. New trees are generated using genetic operators and replace old trees based on a Wilcoxon test using the measured score and tree complexity. According to the authors, this method should be understood as proof of concept and has not yet been used to configure target algorithms with more than one parameter.

\paragraph{PCIT}\label{PCIT}
Similar to ISAC, parallel configuration with instance transfer (PCIT)~\citep{DBLP:conf/aaai/LiuT019} partitions problem instances into clusters and finds a configuration for each cluster. While ISAC produces clusters once, and only afterwards finds a configuration for each cluster, PCIT adjusts clusters and assigns configurations sequentially. This allows PCIT to adjust its clusters according to the configurations for each cluster. This comes at the expense of additional configuration costs.

PCIT employs an instance transfer mechanism to shift around instances between groups as new runtime data becomes available. In particular, this mechanism is used in case it becomes clear that a configurator can not find a configuration that is suitable for all instances within a cluster. The transfer mechanism of PCIT has to (i) identify the instances to transfer and (ii) choose a cluster to transfer the instances to. Instances are selected to be transferred when their runtime using the configuration in the respective cluster is higher than the median runtime of all instances over all clusters. If the instance is to be transferred, a surrgoate is used to predict the runtime of the instance for the configurations of the other clusters. An instance is assigned to the cluster with the lowest predicted runtime. After the transfer, a configurator is run for every cluster. This loop continues until the computational budget is exhausted.  The algorithm configurator SMAC is used to find configurations for each cluster, however any offline configurator can be used. 

\section{Multi-objective Methods}\label{sec:MultiObjective}
We now extend our survey to AC methods that can tune multiple objectives at the same time, such as runtime and quality, or runtime and memory usage~\citep{dang2014motivations}. That is, instead of a single objective, $m(c, \mathcal{I}_\mathit{train}, \vec{\theta})$, multiple objectives $\M:=(m_1,...,m_n)$ may be present and competing with each other. The goal of the configurator is then to find a set of configurations $\hat{\Theta} \subseteq \Theta$ such that no $\vec{\theta} \in \hat{\Theta}$ is dominated by another $\vec{\theta}'$ with respect to some dominance relation $\prec$ over the objective functions~\citep{blot2016mo}. If given preferences from the decision maker, the multiple objectives can be weighted and a single ``best'' configuration $\hat{\vec{\theta}} \subseteq \hat{\Theta} $ can be provided to the decision maker. Note that this AC setting should not be confused with the task of automatically configuring multi-objective target algorithms. The key difference is that the former is concerned with the objective function of the configurator while the latter is interested in the objective function of the target algorithm~\citep{bezerra2020automatic}. Consider that, for example, essentially any single-objective metaheuristic could be configured in a multi-objective way, as quality and runtime present a tradeoff in most of these approaches. The adjustments necessary for single objective configurators to handle multiple configurations are not trivial.

Nevertheless, the multi-objective AC scenario is clearly linked to multi-objective optimization~\citep{coello2006evolutionary}. Multi-objective target algorithms can either be configured by means of multi-objective configurators or by choosing a single objective target metric to be optimized by the configurator, such as the hypervolume or epsilon measure~\citep{lopez2010automatic,lopez2012automatic,blot2017automatically}. In the first case, it is not fully clear how the objective values obtained by running the target algorithm translate into configurator objectives. That is, multi-objective target algorithms usually return a set of solutions that produce different values for the different objectives, which in turn would have to be considered by the configurator.

\begin{figure}[t]
  \centering
  \includegraphics[width=0.95\textwidth]{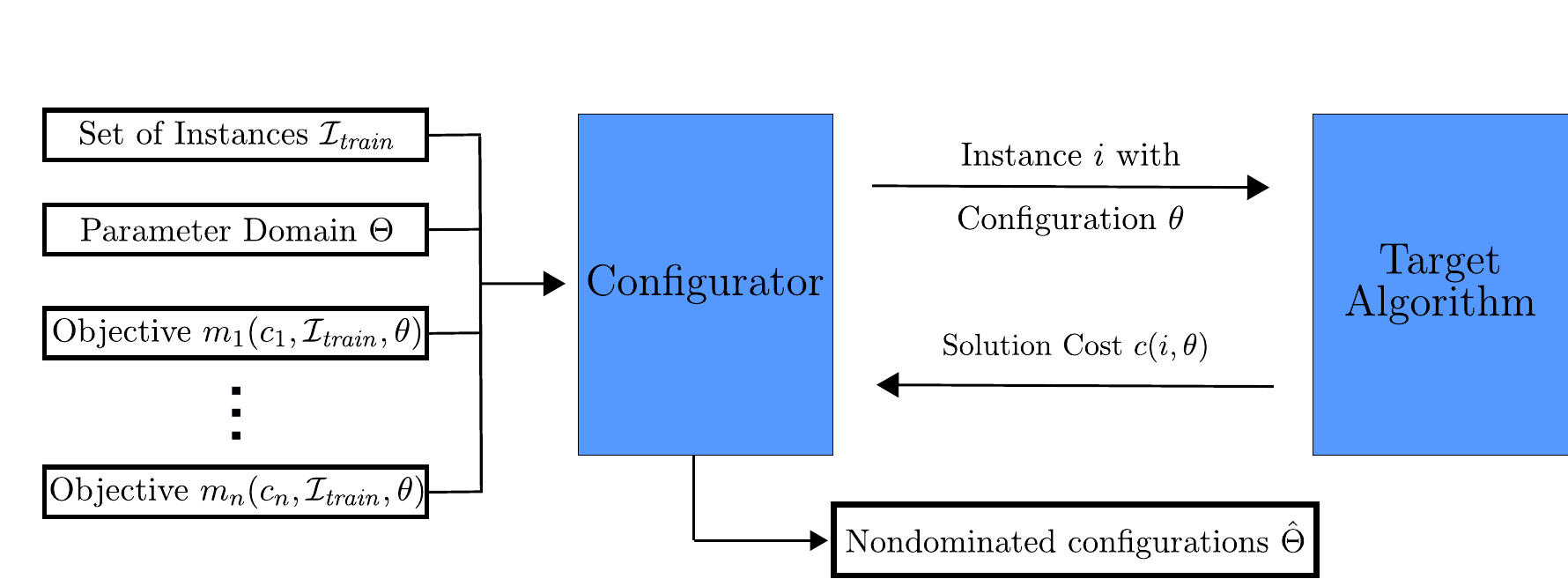}
  \caption{\label{fig:multiobjective} Multi-objective AC}
\end{figure}

\paragraph{MO-ParamILS}\label{MO-ParamILS}
MO-ParamILS~\citep{blot2016mo} is an offline configurator that considers multiple objectives and returns a set of non-dominated configurations for the target algorithm. It is based on the ParamILS framework and, like its predecessor, provides two evaluation techniques while also utilizing the one exchange neighborhood search. In addition, it can run in parallel and can handle large discrete search spaces. 

To handle multiple objectives, MO-ParamILS introduces an archive of configurations and utilizes Pareto dominance to compare configurations. More precisely, it replaces the single configuration that is used during the local search and as incumbent through an archive of non-dominated configurations. In addition, the evaluation mechanism to compare configurations is adjusted to operate on Pareto dominance. In particular, the notion of dominance is used to compare two configurations during local search and also to discard configurations from the archive as new configurations are added.

\paragraph{Multi-objective racing}\label{Multi-ObjectiveRacing}
Racing procedures can be adjusted to take multiple objectives into account. S-Race~\citep{zhang2013s, zhang2015multi} and SPRINT-Race~\citep{zhang2015sprint} are multi-objective configurators similar to F-Race~\citep{birattari2002racing,birattari2009tuning}, which utilize races and statistical testing to find a set of Pareto optimal configurations for discrete parameter spaces. 

S-Race and SPRINT-Race differ from F-Race in terms of the test they apply to discard configurations. Both start with a full factorial design, from which they iteratively eliminate configurations based on the race results. S-Race uses a sign test~\citep{wackerly2014mathematical} to perform a pairwise comparison and eliminate configurations, and Holm's step-down procedure~\citep{holm1979simple} is applied to account for errors. Utilizing the sign test can lead S-Race to unnecessarily often race and compare two configurations between which no dominance relation exists. To save evaluation time, SPRINT-Race~\citep{zhang2015sprint} replaces the sign test with a sequential probability ratio test with indifference zones~\citep{waldsequential}, enabling it to stop running new races when already enough evidence is gathered that a dominance relation is unlikely. Like F-Race, both methods face the drawback of only being able to handle small, discrete configuration spaces. In addition, studies on the competitiveness of the described approaches are limited in scope.

\section{Dynamic Methods}\label{sec:Dynamic}
\newcommand{\instancefeatures}{\mathcal{I}}
\newcommand{\timefeatures}{\mathcal{T}}
\newcommand{\statefeatures}{\mathcal{Q}}
\newcommand{\algorithms}{\mathcal{A}}

Dynamic algorithm configuration \citep{biedenkapp2020dynamic} (DAC) adjusts the configuration \emph{at runtime} instead of committing to a single configuration for solving an entire problem instance.
Thus, rather than recommending a single configuration, DAC approaches generate a policy to adapt the configuration dynamically. Note that even realtime AC commits to a single configuration when running a given instance, while DAC has the freedom to adjust the configuration according to target algorithm behavior during execution. 
Similar to offline AC, DAC can either focus on finding a policy for a set of problem instances or a policy that is tailored towards a single problem instance (i.e., per-instance algorithm configuration).

Two requirements must be met to implement DAC: (1) the algorithm in question needs to support dynamic changes in its configuration and (2) runtime information must be provided to describe the current state of the target algorithm.

DAC approaches consider two different types of features: instance features $\instancefeatures$, which do not change during target algorithm execution, and features encoding the internal state $\statefeatures$ of the algorithm. Examples of state features include the current iteration of a local search algorithm, the current restart number of a SAT method, or the current solution quality for optimization techniques. 

\citet{biedenkapp2020dynamic} provide the first formal definition of the DAC setting, however, there is a significant amount of earlier work for learning dynamic configuration policies \citep{DBLP:conf/icml/LagoudakisL00,DBLP:journals/endm/LagoudakisL01,DBLP:conf/gecco/PettingerE02}.
Such earlier works use the labels parameter control \citep{DBLP:journals/tec/KarafotiasHE15}, online algorithm selection \citep{DBLP:conf/gecco/VermettenRBD19}, adaptive selection/configuration \citep{DBLP:journals/amai/FialhoCSS10,DBLP:conf/ppsn/RijnDB18}, Self-Adaptive Monte Carlo Tree Search~\citep{sironi2018self} or hyper-reactive search \citep{DBLP:conf/aaai/AnsoteguiPST17,DBLP:conf/lion/AnsoteguiHPST18}. For a comprehensive overview of parameter control with respect to parameters of evolutionary algorithms, we refer the interested reader to \cite{DBLP:journals/tec/KarafotiasHE15}.

\citet{DBLP:conf/ppsn/RijnDB18} and \citet{DBLP:conf/gecco/VermettenRBD19} identify potential performance gains through DAC versus static configurations. Based on this insight and data, it is demonstrated in \citet{DBLP:conf/gecco/YeDB21} that performance gains can already be achieved when the algorithm configuration is adapted only once. Furthermore, the hyper-reactive approach of~\citet{DBLP:conf/aaai/AnsoteguiPST17} won several categories at the MaxSAT Evaluation 2016~\citep{maxsateval2016}. Thus, DAC offers significant potential for improving algorithms, however, it does require algorithm designers to more deeply integrate their techniques with AC methods than was performed in the past.
In the following, we discuss the most frequently used approach to DAC, reinforcement learning (RL). While it is the most popular choice, there also exist other approaches such as policy portfolios, autoconstructive evolution, and multi-armed bandits.

\begin{figure}[tb]
  \centering
  \includegraphics[width=0.5\linewidth]{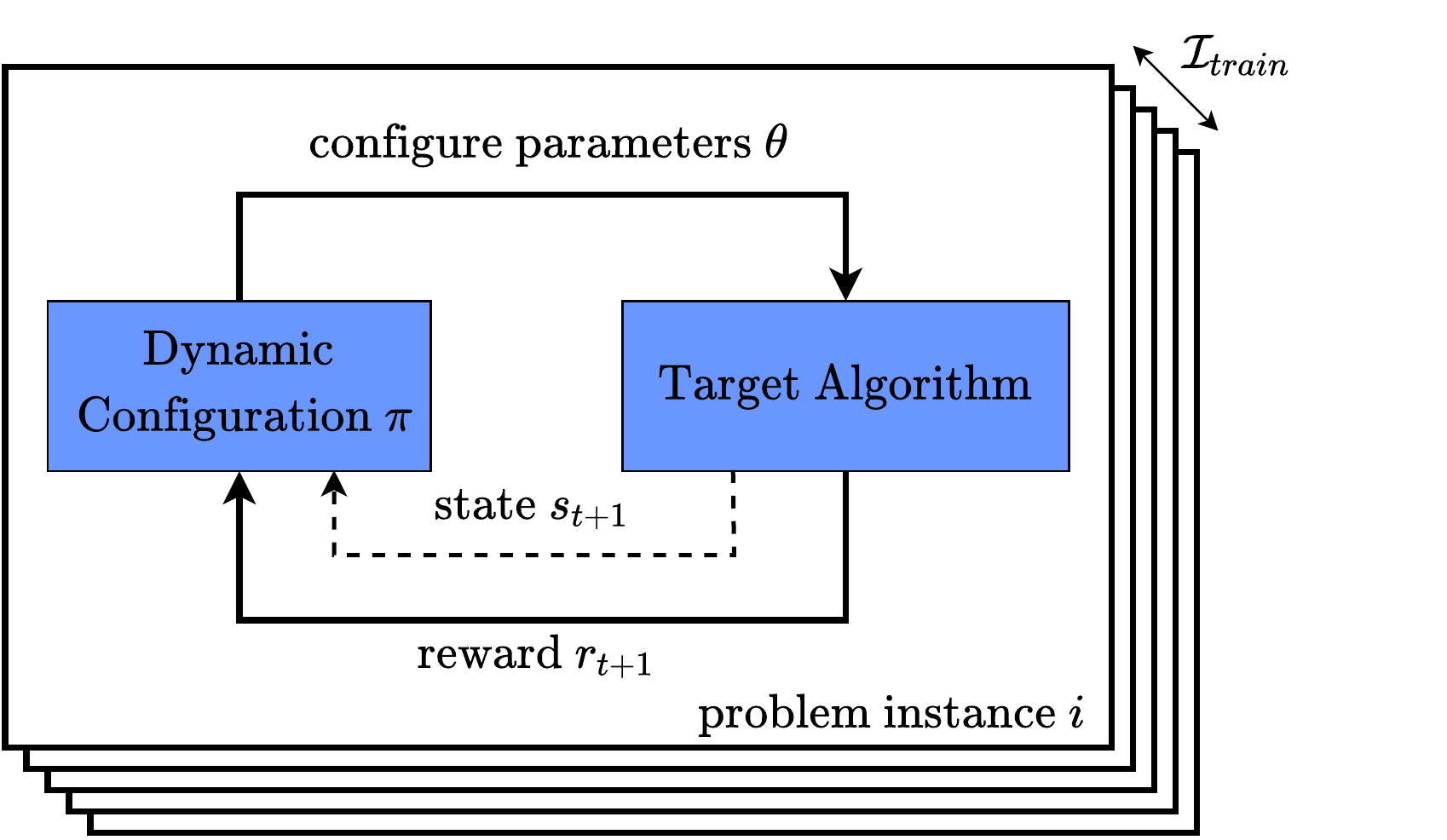}
  \caption{\label{fig:dynamic} Illustration of dynamic AC as presented in~\citep{biedenkapp2020dynamic}}
\end{figure}

\paragraph{Reinforcement learning}\label{p:dac-rl}
A challenge in DAC is to evaluate the quality of individual parameter adjustments since feedback is obtained only at the end of processing a problem instance. Due to this, it is natural to consider DAC within the realm of reinforcement learning (we refer to this as DAC-RL). To this end, the DAC problem is modeled in terms of a Markov decision process (MDP) \citep{DBLP:books/wi/Puterman94}
based on which a policy $\pi: \instancefeatures \times \statefeatures \rightarrow \Theta$ is learned from data \citep{biedenkapp2020dynamic}. As a reward signal, typically, the eventually obtained performance value is propagated back to the agent. Using RL allows a wide variety of researched mechanisms to be directly applied to DAC. 

The roots of DAC extend to a variety of methods that use RL to control parameters of optimization approaches online. For example, in genetic algorithms \citep{sakurai2010method,karafotias2014generic}, planning algorithms \citep{pageau2019configuration,DBLP:conf/aips/0001BHML21,bhatia2021tuning}, hyper-heuristics \citep{DBLP:journals/ijamc-igi/OzcanMOB10}, physics simulations \citep{armstrongWarrenDynAS}, and evolutionary strategies \citep{shala2020learning}. The work of~\citet{sakurai2010method}, \citet{karafotias2014generic} and \citet{DBLP:journals/ijamc-igi/OzcanMOB10} in particular can be widely applied to optimization problems, whereas other works are more application focused.

\paragraph{Hyper-configurable reactive search} \label{p:HCRS}
Under the label hyper-configurable reactive search (HCRS), \citet{DBLP:conf/aaai/AnsoteguiPST17}, \citet{DBLP:conf/lion/AnsoteguiHPST18} and Sellmann and Tierney (\citeyear{DBLP:conf/lion/SellmannT20}) propose an approach to DAC by using standard, offline AC techniques to train a dynamic policy. More specifically, instead of tuning the parameters of an algorithm directly, each parameter is determined by a logistic regression that accepts runtime features from the target algorithm. The parameters of the logistic regressions are exposed to the offline AC technique, GGA++. The target algorithm can thus cheaply query the values it needs, when it needs them, according to the current search state. In~\citet{DBLP:conf/aaai/AnsoteguiPST17}, the problem instances are further grouped via the CSHC algorithm~\citep{malitsky2013algorithm} and for each of the resulting clusters, such a policy encoded via logistic regression models is configured according to the ISAC method~\citep{kadioglu2010isac}.

\paragraph{Autoconstructive evolution}
In autoconstructive evolution, the aim is to not only evolve a solution to an optimization problem, but also to evolve the genetic operators that are used for recombination, selection, and diversification \citep{DBLP:journals/gpem/SpectorR02,DBLP:conf/gecco/HarringtonSPO12,DBLP:conf/gecco/SpectorM17a}. To this end, each individual represents a tuple of a solution encoding and the encoding of the programs for the genetic operators, respectively. Typically, genetic programming is applied to this problem setting since it can represent programs naturally in terms of abstract syntax trees. In contrast to other algorithm configuration approaches, considering only Boolean, numeric, or nominal parameters of algorithms, the configuration of such genetic operators is even more challenging. Additionally, autoconstructive evolution goes beyond what is usually done in DAC. It adapts the genetic operators for each individual of the population so that the mating behavior changes over time and is thus specific for each individual. Another difference is that for adapting the genetic operators of the evolutionary algorithms, there is no need to engineer and monitor any features representing the algorithm's state. Although Spector and Moscovici (\citeyear{DBLP:conf/gecco/SpectorM17a}) obtain promising initial results, the combined evolution of solutions and their genetic operators is comparatively challenging, and it remains an open question whether autoconstructive evolution is indeed superior to non-autoconstructive evolution.

\paragraph{Multi-armed bandits}
DAC can be cast as a multi-armed bandit problem, where each configuration represents an arm with an associated reward distribution. The agent is then tasked to maximize the cumulative reward, i.e., ``pull the arm`` that maximizes the reward. Put differently, the agent aims to select a configuration that works well for the current time step. This setting of the bandit problem is particularly challenging because the reward distributions associated with the respective arms change over time. In \citet{DBLP:journals/amai/FialhoCSS10}, various multi-armed bandit approaches are compared with an approach that only considers rewards obtained within a certain time frame (i.e., a sliding window) to allow for smoother transitions between configurations. Note that in this work, the authors assume rewards to be retrievable during an algorithm run. While this assumption holds for many optimization approaches, it does not necessarily hold for all of them (e.g., the pre-processing/root node phase of solving mixed-integer programs), and also often does not hold for simulations.

\section{Research Directions}\label{ResearchDirections}

Many fruitful avenues of research remain in the area of AC. We formulate research directions with three themes in mind. First, we consider what is required for AC techniques to be more widely adopted in industry. Second, we investigate algorithm configuration settings that have not yet been considered. Third, we consider what methodological advances could move the field forward.

\subsection{Industry Adoption}

Compared to AC, HPO has begun to be adopted in both industry and academia~\citep{van2021automl}. Furthermore, HPO is the focus of a number of start-ups, such as SigOpt (which was acquired by Intel~\citep{sigopt}), MindsDB, pecan.ai, and dotData, among many more. AC, however, has received comparatively little fanfare or adoption. While some well-known AC tools have public releases, such as SMAC~\citep{lindauer2021smac3}, irace~\citep{lopez2016irace}, and GGA~\citep{ansotegui2009gender}, we are unaware of widespread use of these in industry. The company Optano GmbH released a version of the GGA configurator~\citep{optano}\footnote{The configurator from Optano GmbH is the result of a joint research project between Optano GmbH and Kevin Tierney, but we note that it is freely available and there is no commercial interest on the part of Kevin Tierney.}, making it perhaps the first commercially developed, general-purpose AC software. The mixed-integer programming solvers Gurobi and CPLEX both contain tools to adjust their parameters. However, to the best of our knowledge, these tools are both outperformed by publicly available research tools. Furthermore, we do not know how either of these tools work.

We speculate that there are several reasons these tools are not frequently used in industry. First, the target algorithm must be carefully designed with its parameters exposed through a command-line interface as proposed by the programming by optimization paradigm~\citep{pbo}. 
Second, in some cases, users may not think they have sufficient data available to properly configure their approach. 
This is related to the \textit{cold-start problem} in ML, in that users of AC systems may not have gathered enough instances to properly train their parameterized algorithm to solve their problem effectively. While some work has been performed in this direction in terms of automatically generating instances~\cite{malitsky2016structure}; Smith-Miles and Bowly (\citeyear{smith2015generating}); \cite{akgun2019instance,DBLP:journals/tec/TangLYY21,DBLP:journals/tcyb/LiuTY22}, there are still many real-world problems that cannot be modeled with these techniques.

Third, the runtimes of some target algorithms are too high for AC. In these cases, transfer learning from a dataset of similar, but easier, problem instances may be a way forward. Alternatively, the AC algorithms could try to guess the quality of the target algorithm before it is finished, or at least guess a ranking over the different configurations.

A fourth reason for a lack of adoption in industry is likely the inability of current algorithm configuration tools to integrate into existing systems. Target algorithms may not be easily accessible over the command line or accept data in a simple ``problem instance'' format as required by existing configurators. For example, if an algorithm is tightly integrated with a database or SAP system, AC may require significant extra work. The solution here is likely not a new research concept or method, but rather raising awareness that decoupling solvers from production environments will enable configuration of the algorithm's parameters.

A final, fifth reason is the expert knowledge required to set up an AC environment. AC is not a topic that is widely taught in data science or computer science curriculums, although at least HPO is beginning to be adopted. Nonetheless, a deep familiarity with the various components of AC is necessary to successfully implement AC in practice. Furthermore, we are not aware of many consultants offering AC services to fill this knowledge gap.

\subsection{Novel AC Settings}

We identify several AC settings according to our problem view that have not yet been realized in the literature. In particular, settings considering multiple objectives, especially in terms of the target algorithm, during run target algorithm observations, dynamic configuration adjustment, and parameter transfer learning have not been significantly considered in the literature. It is easy to imagine real-world applications for all the situations listed, thus there remain ample research opportunities available in AC.

\paragraph{Multiple objectives}
In Section~\ref{sec:MultiObjective}, we identify several approaches for multi-objective AC, but note that these are rather limited in how they incorporate multiple objectives into their configuration. Especially in regard to runtime objectives, multi-objective AC methods are still lacking adaptive mechanisms for capping runs or generating configurations that target specific areas of the Pareto front. While we expect multi-objective target algorithms to integrate into these multi-objective AC techniques, extra heuristics or care could be taken to consider the Pareto front of the target algorithm in addition to (or instead of) a Pareto front over configurations. Furthermore, multiple objectives have only been included in offline configuration. 

Since realtime AC is tasked with returning solutions directly to users, providing multiple solutions in a multi-objective context could be valuable. Realtime configurators only get one shot at solving a problem instance (or, in this case, generating multiple solutions). Thus, the research focus would be on generating an interesting set of solutions for the user. This set ought to adequately represent the Pareto front while providing the solutions quickly if runtime is one of the considered objectives.

\paragraph{Incorporating runtime information \& DAC}

While adjusting algorithm configurations dynamically is not a new concept, learning policies with AC techniques is, and as pointed out in Section~\ref{sec:Dynamic}, has shown considerable promise. These techniques use information from the target algorithm about its current state to help create an effective policy for setting parameters. Many algorithms offer indications as to how good (or bad) their solution procedure is progressing, especially when compared to the output of other configurations solving the same instance. However, little work has been done on using this information in the AC approach itself, e.g., to stop unpromising configurations early or to assist in comparing configurations with censored runtime data. This line of work has the potential to reduce the overall computation time of AC.

\paragraph{Meta-AC} 

AC methods themselves contain many parameters that require tuning, but, in a twist of irony, the computational cost of configuring a configurator makes this a daunting task. The AS problem has been similarly investigated on a meta-level by \citet{tornede2020towards,tornede2021algorithm} for learning ensembles of algorithm selectors. Moreover, this sort of ``meta configuration'' has been performed by~\citet{ansotegui2018} to configure algorithm selection portfolios with GGA and \citet{lindauerHHS17}. Given that considering the ``hyper''-hyperparameters in HPO has seen also some success~\citep{feurer2018towards}, we see potential for the general AC setting as well.

\paragraph{Transfer learning} 

Consider a target algorithm for which a new version is released with several new parameters, and perhaps some existing parameters have different domains or are removed. From a practical standpoint, there are ways of handling this in modern AC software, such as inserting the previously tuned parameters along with some default values for the rest as a starting point of the search. Of course, this involves starting a brand-new configuration process that ``forgets'' everything that was previously learned about how the previous parameters behaved on a set of instances. Maintaining this state and transferring knowledge to the new or modified parameters could result in better configurations and less computation time used on AC.
A first step towards this was recently proposed by Franzin and Stützle (\citeyear{franzin2020towards}).

\paragraph{Generalization} 
Configurations found by offline configurators (see Section \ref{sec:ModelFree} and \ref{sec:ModelBased}) might not always perform as well in production as one might expect based on the achieved training performance. That is, the configurations found may not generalize well to new instances or performance may degrade over time due to concept drift. To address this, the community has shifted focus towards methods that allow for an active control of the configuration during production based on the new instance, e.g., in the realtime and dynamic configuration settings. Nonetheless, more attention is needed to ensuring that configurations remain effective in production settings.

\subsection{Novel Benchmarks}

To accelerate future research and to provide reproducible results, benchmarks for new problem settings are needed. AClib~\citep{hutter2014aclib}, the currently most widely used benchmark library in the context of AC, covers the offline configuration case and provides a set of different problems (SAT, MIP, ASP, etc.) of varying complexity (number of variables and problem instances) for tasks of runtime or quality configuration. For DAC, the DACBench has been proposed~\citep{eimer2021dacbench}, although this does not support DAC settings envisioned, e.g., by hyper-reactive search. As an alternative to such libraries, AC methods can also be benchmarked by using surrogate models that are trained on test instances in advance, resulting in cheaper evaluations when testing~\citep{eggensperger2018efficient}. The existing benchmarks fail to cover other configurations settings like the realtime configuration setting or the configuration of multi-objective target algorithms.

\subsection{Novel AC Methodologies}

AC methods have become extremely sophisticated and cover a wide range of methodologies including evolutionary algorithms, statistical tests, and learned surrogate models. There nonetheless remain opportunities to improve current methods and create better AC algorithms. We note that our goal in this section is not necessarily to specify the methodologies of the future, but rather to identify the key challenges that remain in the hopes that engaged readers will fill these gaps with their ideas. To this end, we discuss several challenges faced by AC approaches: comparing/ranking configurations, generating new configurations, and selecting instances.

\paragraph{Comparing/ranking configurations} This challenge can be summarized as follows: given two or more configurations, determine which one(s) is(are) the best performing without needing to run the configurations on the entire training set of instances. In offline configuration, the ranking can be used to generate new configurations (part of the next challenge), whereas in realtime configuration the rankings can be critical to deciding which configurations are allowed to try to solve the current instance. Methods for ranking and comparing include using empirical hardness models~\cite{leyton2009empirical} or more general surrogates as in GGA++~\citep{ansotegui15gga++} or SMAC~\citep{hutter2011sequential}, statistics~\citep{lopez2016irace}, the TrueSkill mechanism~\citep{fitzgerald2015reactr}, bandits~\citep{el2020pool}. Nonetheless, there is undoubtedly still room for improvement, using perhaps new preference learning techniques (as have been used for AS in~\citet{hanselle20hybrid}) or deep learning models.

\paragraph{Generating new configurations}

Every AC algorithm must have a mechanism for generating new configurations. The question, of course, is how to generate configurations that are high quality with respect to the configuration objective. This research question is closely related to comparing and ranking configurations, as mechanisms like the one in GGA++ compare configurations that are generated according to some rules. Creating new configurations, especially in an instance-specific capacity, offers a clear path to higher quality AC mechanisms that can be used across a wide range of AC problem settings.

\paragraph{Instance selection}

This review begins by identifying instance handling mechanisms as one of the key differences between HPO and general AC. The strategies used in the literature are fairly basic. Most algorithms either select random subsets of the instances or expand a subset. It might be worthwhile to adapt methods related to the concept of dataset distillation \citep{wang2018dataset} to generate small instance sets that are nonetheless representative of the complete instance set to speed up configuration.
Instance-specific approaches usually find some way to partition the instances before tuning multiple configurations. However, much remains unknown about how certain orders of considering instances could change or improve the configurations found. It is possible that the order in which instances are examined does not matter, but given that this is a core competency of general AC methods, the potential for performance improvements is present.

\paragraph{Surrogate model features}
Learning suitable features for surrogate models automatically is closely related to instance selection. Even though there are good features for particular problem classes such as SAT or MILP problems~\cite{hutter2014algorithm}, other not so well studied problems lack well-defined features and require domain experts to manually derive them for a specific use case. Previous works~\citep{kroer2011feature,tierneyalgorithm} have begun to address this problem in the realm of AS. Adopting this for AC, however, remains an open challenge.

\paragraph{Bounded rationality and rational metareasoning}
\citet{hullermeier2021automated} elaborate on automated machine learning (as well as related problems of automated algorithm design, such as hyperparameter optimization and algorithm selection) from the perspective of bounded rationality. The authors propose to view methods for automated algorithm design as intelligent agents, which need to take decisions under bounded resources (e.g., configuration time in the case of AC) and thus apply reasoning on a meta-level to decide how to optimally allocate the available resources. The motivation for adopting this perspective is twofold: first, to shed light on existing AutoML methods, and second, to inspire new approaches based on established (decision-theoretic) principles of rational metareasoning and bounded optimality \citep{russ_dt,russ_ra97,zilb_ma08,russ_ra16}.

\subsection{Theoretical Results}
Although the field of AC has seen significant progress recently in terms of answering theoretical questions, there are still a number of critical challenges that remain. 
In the following, we address some of these challenges.

\paragraph{Generalization guarantees}
The recent work by Pushak and Hoos (\citeyear{pushak2018algorithm}) shows empirically that the configuration landscape, i.e., the class of cost functions, exhibits an approximately uni-modal or convex structure for a couple of NP-hard problem classes and sophisticated target algorithms.
In light of this, it seems reasonable to extend the current results regarding generalization guarantees to the scenario, where the functions in the class of cost functions are (approximately) uni-modal or convex.
Such an assumption will likely lead to better theoretical generalization guarantees as the current ones reviewed in Section \ref{subsec:gen_guarantees}, as uni-modalities and/or convexity of target functions are known to accelerate the learning process in general.

\paragraph{Theoretical analysis of state-of-the-art configurators}
Although the work of \cite{Hall2019Cutoff} provides valuable insights into the theoretical properties of ParamILS, it still does not cover all its theoretical properties.
Even less is known about the theoretical properties of other state-of-the-art algorithm configurators, such as SMAC, irace, or GGA, besides that SMAC will find the optimal configuration if the configuration space is finite and sufficient time for running SMAC is available (see Theorem 4 in \cite{hutter2010sequential}).
A deeper theoretical analysis of these algorithms is likely to be helpful to understand the reasons for the practically appealing behavior of these configurators. Furthermore, this analysis could reveal on which classes of problems the configurators are likely to perform effectively, and on which they are not.

\paragraph{Instance specific theoretical analyses}
So far, the existing theoretical approaches do not take additional side information in the form of feature vectors of the problem instances into account, i.e., they all employ a one-fits-all paradigm.
However, as several of the reviewed works in Section \ref{sec:Instancespecific} have shown, there might be a benefit regarding the performance of an algorithm configurator in practical applications if these features are incorporated.
In light of this, it would be interesting to investigate theoretically the potential improvement of an instance-specific (i.e., feature-based) configurator over a one-fits-all paradigm (i.e., feature-free) configurator for specific problem scenarios.

\section{Conclusion}\label{sec:conclusion}
Parameters are ubiquitous in modern optimization approaches and beyond, with all of the significant solvers for, e.g., MILP, SAT, or TSP problems containing parameters that influence their performance and need to be set by the user. AC frees the user from this tedious and error-prone task by automating the search for high-quality configurations. This survey presented an overview of the current state of AC research by outlining relevant methods, their design features, and problem settings. In particular, we provided two taxonomies for organizing AC approaches, reviewed and contrasted different approaches in the light of the problem setting they were designed for, and highlight underlying principles and ideas. Overall, we find that the AC literature is in a mature state, including powerful empirical approaches available to handle a variety of large-scale, real-world challenges, as well as theoretical approaches providing quality guarantees and an understanding of the AC domain.

We identify that the methodological trend is towards incorporating learned models into AC methods and that this has improved performance in every AC setting where it has been attempted. Nonetheless, the AC literature shows a surprising amount of hybridization of local search, evolutionary and model-based methods. We hypothesize that there is still significant progress that can be made in the area of AC, despite the sophistication of current methods, and are encouraged by the significant increase in attention the field has received, in particular through the spread of HPO techniques. Finally, we especially encourage researchers to address the real-world usability of AC techniques to ensure that the promising performance gains the AC community is seeing can benefit the world at large.

\section*{Acknowledgements}
This work was partially supported by the German Research Foundation (DFG) within the Collaborative Research Center ``On-The-Fly Computing'' (SFB 901/3 project no.\ 160364472) and by the research training group ``Dataninja'' (Trustworthy AI for Seamless Problem Solving: Next Generation Intelligence Joins Robust Data Analysis) funded by the German federal state of North Rhine-Westphalia. We also would like to thank the anonymous reviewers for their suggestions on the first version of this manuscript.

\section{Appendix}
To help the reader navigate though the jungle of AC, we provide additional resources. Table \ref{table:abrevations} contains a list of abbreviations with terms related to AC used within this work. In addtion, we provide a list of software resources (Table \ref{table:usefullsoftware}) that contains currently available tools for AC. We only include software that is widely used.

\begin{table}[!hp]
\begin{tabularx}{\columnwidth}{lX} 
    \hline
	\multicolumn{2}{c}{\textbf{General}} \\
	\hline
	AC & Algorithm configuration \\
	AS & Algorithm selection \\
	BN & Bayesian networks\\
	BO & Bayesian optimization \\
    CASH & Combined algorithm selection and hyperparameter optimization\\
    DAC & Model-based algorithm tuning engine \\
    DOE & Design of experiments methodology\\
    EA & Evolutionary algorithm\\
    EI & Expected improvement \\
    GP & Gaussian processe\\
    HPO & Hyperparameter optimization \\
    HCRS & Hyper-configurable reactive search \\
    ILS & Iterated local search \\
    MILP & Mixed-integer programming\\
    RVNS & Reduced variable neighborhood search \\ 
    SAT & Boolean satisfiability problems\\
    SMBO & Sequential model-based optimization\\
    UCB & Upper confidence bound\\
    
    \hline
	\multicolumn{2}{c}{\textbf{Method related}} \\
	\hline
	BNT & Bayesian network tuning \\
	CAR & CapsAndRuns \\
	CPPL & Contextual preselection with Plackett-Luce \\
	D-SMAC & Distributed SMAC \\
	EISAC & Evolving instance-specific algorithm configuration \\
	GGA &Gender-based genetic algorithm \\
	GPS & Golden Parameter Search \\
    HORA & Heuristic oriented racing algorithm\\
    ICAR & ImpatientCapsAndRuns \\
	I/F-Race & Iterated F-Race \\
	ISAC & Instance-specific algorithm configuration \\
	MATE & Model-based algorithm tuning engine \\
	MBGM & Model-based graphical methods\\
	ReACT & Realtime algorithm configuration through tournaments \\
	REVAC & Parameter relevance estimation and value calibration \\
	ROAR & Random online aggressive racing \\
	SKO & sequential kriging meta-modelling \\
	SMAC+PS & Sequential model-based optimization for algorithm configuration\\
	SMAC & SMAC and probabilistic sampling\\
	SPO & Sequential parameter optimization\\
	SP & Structured procrastination \\
	TB-SPO & Time-bounded SPO \\
\end{tabularx} 
\caption{Glossary of acronyms used in this work}
\label{table:abrevations}
\end{table}

\begin{table}[!ht]
\begin{tabularx}{\columnwidth}{lX} 
    \hline
	\multicolumn{2}{c}{\textbf{General AC systems}} \\
	\hline
	D-SMAC &  \href{https://github.com/tqichun/distributed-SMAC3}{https://github.com/tqichun/distributed-SMAC3}\\
	GPS &  \href{https://github.com/YashaPushak/GPS}{https://github.com/YashaPushak/GPS}\\
	irace &  \href{https://github.com/MLopez-Ibanez/irace}{https://github.com/MLopez-Ibanez/irace}\\
    OAT (GGA) &  \href{https://docs.optano.com/algorithm.tuner/current/}{https://docs.optano.com/algorithm.tuner/current/}\\
	ParamILS & \href{https://www.cs.ubc.ca/labs/algorithms/Projects/ParamILS/}{https://www.cs.ubc.ca/labs/algorithms/Projects/ParamILS/}\\
    PyDGGA &  \href{http://ulog.udl.cat/software/}{http://ulog.udl.cat/software/}\\
	REVAC & \href{https://github.com/ChrisTimperley/RubyREVAC}{https://github.com/ChrisTimperley/RubyREVAC}\\
    SMAC 3 &  \href{https://github.com/automl/SMAC3}{https://github.com/automl/SMAC3}\\
    \hline
    
    \hline
	\multicolumn{2}{c}{\textbf{Benchmarks}} \\
	\hline
	AClib & \href{https://bitbucket.org/mlindauer/aclib2/src/master/}{https://bitbucket.org/mlindauer/aclib2/src/master/}\\
	DAC & \href{https://github.com/automl/DAC}{https://github.com/automl/DAC}\\
\end{tabularx}
\caption{List of available software in the realm of AC.}
\label{table:usefullsoftware}
\end{table}

\newpage
\bibliography{literature/dynamic_algorithm_configuration,literature/benchmarks, literature/instance_specific, literature/model_based, literature/model_free_methods, literature/multi-objective, literature/problem_formulation, literature/realtime, literature/theoretical_guarantees, literature/classification, literature/introduction,
literature/research_directions}

 
\end{document}